\documentclass[]{x2lab}
\usepackage{fix-cm}
\usepackage{csquotes}
\usepackage{afterpage}
\usepackage{amsmath,amssymb,amsfonts}
\usepackage{mathtools}
\usepackage{amsthm}
\usepackage{xspace}
\usepackage{colortbl}
\usepackage{enumitem}
\usepackage{wrapfig}
\usepackage{makecell}
\usepackage{pifont}
\usepackage{svg}
\usepackage{array}
\usepackage{tabularx}
\usepackage{longtable}
\usepackage{xltabular}
\usepackage{float}
\usepackage{placeins}

\graphicspath{{figures/}{figs/}}

\definecolor{fbApp}{HTML}{c8e7fa}
\definecolor{fbPurple3}{HTML}{f0ebf5}
\definecolor{citecolor}{HTML}{0071BC}
\definecolor{linkcolor}{HTML}{ED1C24}

\titleformat*{\paragraph}{\rmfamily\bfseries}

\title{X-Tokenizer: A Multimodal Action Tokenizer for Vision-Language-Action Pretraining}

\author[1,2,3,*]{Miracle Kang}
\author[1,*\dagger]{Lights Shi}
\author[1,\dagger]{Lucy Liang}
\author[1]{Roy Gan}
\author[1,3]{Dongxiu Liu}
\author[1]{Pushi Zhang}
\author[1]{Sylas Chen}
\author[1]{Shawn Qin}
\author[3]{Yinan Zheng}
\author[3]{Jinliang Zheng}
\author[1]{Hao Wang}
\author[3,\ddagger]{Xianyuan Zhan}
\author[1,3,\ddagger]{Hang Su}

\affiliation[1]{\textbf{X SQUARE ROBOT}}
\affiliation[2]{\textbf{City University of Hong Kong}}
\affiliation[3]{\textbf{Tsinghua University}}
\contribution[*]{Equal Contributions}
\contribution[\dagger]{Project Lead}
\contribution[\ddagger]{Correspondence Authors}

\abstract{Modern Vision-Language-Action (VLA) models must bridge pretrained
vision-language reasoning and precise continuous robot control. Existing
action tokenizers discretize actions primarily for reconstruction, producing
codes that preserve motion geometry but provide only weak semantic supervision
to the backbone. We therefore formulate action tokenization not as mere
compression, but as \emph{semantic interface learning} between multimodal
reasoning and executable control. To this end, we introduce
\textbf{X-Tokenizer}, a lightweight encoder--Semantic Residual Quantization
(SRQ)--decoder architecture that provides a shared action interface across
diverse robotic arm embodiments. Its key component, \textbf{SRQ}, imposes an
asymmetric structure on residual vector quantization: the first level is
trained with \textbf{Masked Action Modeling (MAM)} to form a discrete
\emph{action language} that captures coarse motion intent, while deeper
levels remain reconstruction-oriented residuals that preserve fine-grained
details. To further align action tokens with multimodal semantics,
X-Tokenizer is pretrained with contrastive alignment to the representation
space of a pretrained foundation model and with next-frame vision-language
feature prediction. Pretrained on \textbf{2.4M trajectories (2.0B action
frames)}, a single frozen X-Tokenizer plugs into a mixed discrete-continuous
VLA as a representation-shaping supervision signal. X-Tokenizer achieves top
real-world aggregate and strong RoboTwin 2.0 simulation results.
Outperforming FAST in multimodal grounding ($+13.5\%$) and long-horizon
tasks ($+8.25$), it shows that action tokenizers serve as semantic interfaces
for VLA pretraining beyond mere action compression.\\}

\date{\textbf{June 2026}}

\metadata[Project Page]{\url{https://x-square-robot.github.io/X-Tokenizer_projectPage/}}

\begin{document}
\maketitle

\section{Introduction}
\label{sec:intro}

Modern embodied foundation models---including Vision-Language-Action (VLA)
policies with continuous action heads~\cite{bjorck2025gr00t,zheng2025x,black2024pi_0,chi2025diffusion,li2025controlvla},
latent world models~\cite{assran2025v,bi2025motus,liu2025efficient,maes2026leworldmodel}, and video-generation
policies~\cite{cen2025worldvla,li2026wallwmcarvingworldaction,li2026causal}---increasingly seek to couple pretrained
multimodal reasoning with executable robot control. A central difficulty lies
in the representation mismatch between these two regimes: pretrained
vision-language backbones operate over semantically structured discrete
representations, whereas robot policies must ultimately produce precise,
continuous motor commands. Action tokenizers provide one route for bridging
this gap by mapping continuous action chunks into discrete symbols that can be
predicted by a VLM-style backbone. Yet most existing tokenizers are designed
primarily as compression modules: they minimize reconstruction error under a
fixed token budget and consequently produce codes that partition the geometric
action space, but are not explicitly aligned with task semantics, visual
context, or language-conditioned intent.

This limitation is especially consequential in hybrid
discrete-continuous VLA systems~\cite{black2025pi_,intelligence2026pi07steerablegeneralistrobotic,yu2026wall,zhai2025igniting,liu2025hybridvla,zheng2025universal,jiang2025galaxea,wu2026pragmatic}: the discrete
action-token prediction loss is not merely an auxiliary objective but
also shapes the shared hidden states on which a downstream continuous
expert relies. If the token targets are arbitrary reconstruction
indices, the autoregressive loss weakly supervises the pretrained VLM,
pulling its hidden states toward geometric code patterns rather than
action-relevant multimodal semantics. We therefore formulate action
tokenization as \emph{semantic interface learning}: action tokens
should serve as representation-shaping targets that connect high-level
vision-language reasoning with executable continuous control. Under
this view, a useful tokenizer should satisfy two requirements. First,
its discrete codes should be semantically aligned with the pretrained
backbone so that autoregressive token prediction preserves, rather than
erodes, multimodal grounding. Second, it should still retain sufficient
low-level detail to reconstruct precise robot actions.

Existing action tokenizers only partially satisfy these requirements.
Reconstruction-oriented methods such as FAST~\cite{pertsch2025fast},
VQ-BeT~\cite{lee2024behavior}, VQ-VLA~\cite{wang2025vq}, and
FASTer~\cite{liu2025faster} produce compact action codes with strong signal
fidelity, but they are not explicitly optimized to align their token structure
with visual-language representations. ActionCodec~\cite{dong2026actioncodec}
moves toward cross-modal action representation by introducing contrastive
supervision~\cite{li2024decisionnce,radford2021learning,he2020momentum}, but its alignment is not directly anchored to a frozen pretrained
VLM representation space, and it does not explicitly separate semantic intent
from residual execution detail along the depth of the quantizer.

To instantiate semantic interface learning, we introduce \textbf{X-Tokenizer},
a lightweight cross-embodiment action tokenizer with an
\emph{Encoder--Semantic Residual Quantization (SRQ)--Decoder} architecture,
pretrained on \textbf{2.4M trajectories comprising 2.0B action frames}
across $17$ arm families. Its core design,
\textbf{Semantic Residual Quantization (SRQ)}, imposes an asymmetric structure
on residual vector quantization~\cite{lee2022autoregressive}: the first RVQ
level is trained with \textbf{Masked Action Modeling (MAM)}~\cite{devlin2019bert},
a masked-prediction objective over action tokens, to form a discrete
\emph{action language} that captures coarse motion intent, while deeper RVQ
levels remain reconstruction-oriented residuals that preserve fine-grained
execution details. To further inject multimodal semantics, X-Tokenizer uses
two additional pretraining signals: contrastive alignment to the
representation space of a frozen foundation model and prediction of next-frame vision-language features. These auxiliary heads are used only during tokenizer pretraining and removed afterward, so they introduce no online visual-feature extraction or dynamics-rollout cost. During downstream VLA co-training, the frozen X-Tokenizer provides multi-level action tokens as autoregressive supervision, acting as a semantic scaffold for the VLM backbone while a continuous Flow Matching expert is conditioned on the resulting hidden states to regress executable action trajectories.

Empirically, X-Tokenizer improves the coupling between multimodal grounding
and continuous control. Across RoboTwin~2.0 simulation, real-robot tabletop
tasks, and multimodal VQA evaluation, it achieves strong simulation
performance and the best real-world aggregate among the evaluated action
interfaces. Compared with the reconstruction-only tokenizer FAST,
X-Tokenizer improves multimodal grounding by a $\mathbf{+13.5\%}$ relative
margin ($75.7\!\to\!85.9$) and long-horizon task performance by a
$\mathbf{+8.25}$ absolute margin ($61.0\!\to\!69.25$). These results support
the view that action tokenizers can function as reusable semantic interfaces
for VLA pretraining, rather than merely as internal action-compression
modules.

\section{Related Work}
\label{sec:related}

\subsection{Action Space Design in Robotic Foundation Models}

Modern Vision-Language-Action (VLA) models adopt different action-space
parameterizations, trading off semantic grounding and control fidelity.
\textbf{Discrete autoregressive heads}~\cite{zitkovich2023rt,kim2024openvla}
inherit the token-level modeling interface of pretrained VLMs, but require long
action-token sequences and suffer from discretization error. \textbf{Continuous
generative heads}~\cite{chi2025diffusion,zhao2023learning,black2024pi_0}
preserve smooth, fine-grained control, but their regression or generative
objectives are less directly aligned with language-token training.
\textbf{Hybrid heads}~\cite{black2025pi_,liu2025hybridvla} combine both
interfaces, yet their benefit depends on whether the discrete branch provides
semantically meaningful supervision rather than arbitrary action indices.
This motivates a structured \emph{action tokenizer}: an interface that converts
continuous actions into discrete supervisory signals capable of shaping
multimodal representations while retaining sufficient information for precise
continuous control.

\subsection{Action Tokenizers for Robotics Foundation Models}

Existing action tokenizers mainly differ in how their codebooks are supervised.
Reconstruction-oriented methods, such as FAST~\cite{pertsch2025fast},
VQ-BeT~\cite{lee2024behavior}, VQ-VLA~\cite{wang2025vq},
FASTer~\cite{liu2025faster}, and OAT~\cite{liu2026orderedactiontokenization},
treat tokenization primarily as action compression. While they preserve
trajectory geometry, their tokens are not explicitly aligned with visual
context, language-conditioned intent, or task semantics, making them less
effective as supervision for pretrained multimodal backbones. ActionCodec
\cite{dong2026actioncodec} introduces cross-modal contrastive supervision, but
its alignment space is learned internally rather than anchored to a frozen
pretrained VLM, and its hierarchy does not explicitly separate semantic intent
from execution residuals. Concurrent work CLAP~\cite{clap2026} similarly uses
contrastive alignment to bridge action and visual latents, but operates on
visual dynamics features. Another concurrent route, UniT~\cite{unit2026}, jointly embeds action, visual, and fused  features into a unified codebook. However, such a coupled formulation enforces tight multimodal dependencies during inference; mapping an action trajectory to its discrete codes requires both the images and actions to pass through a computationally heavy multi-stream encoder, which limits its flexibility as a plug-and-play action interface.\\

We formulate action tokenization as \emph{semantic interface learning}.
X-Tokenizer combines asymmetric residual quantization, frozen-VLM contrastive
alignment, and next-frame vision-language feature prediction, so the first RVQ
level captures semantic action intent while deeper levels encode residual
control detail. These objectives are used only during tokenizer pretraining and
removed at deployment, yielding a lightweight interface that regularizes VLA
hidden states without additional perception or dynamics-model cost.

\section{Method}
\label{sec:method}

\begin{figure}[t]
    \centering
    \includegraphics[width=\linewidth]{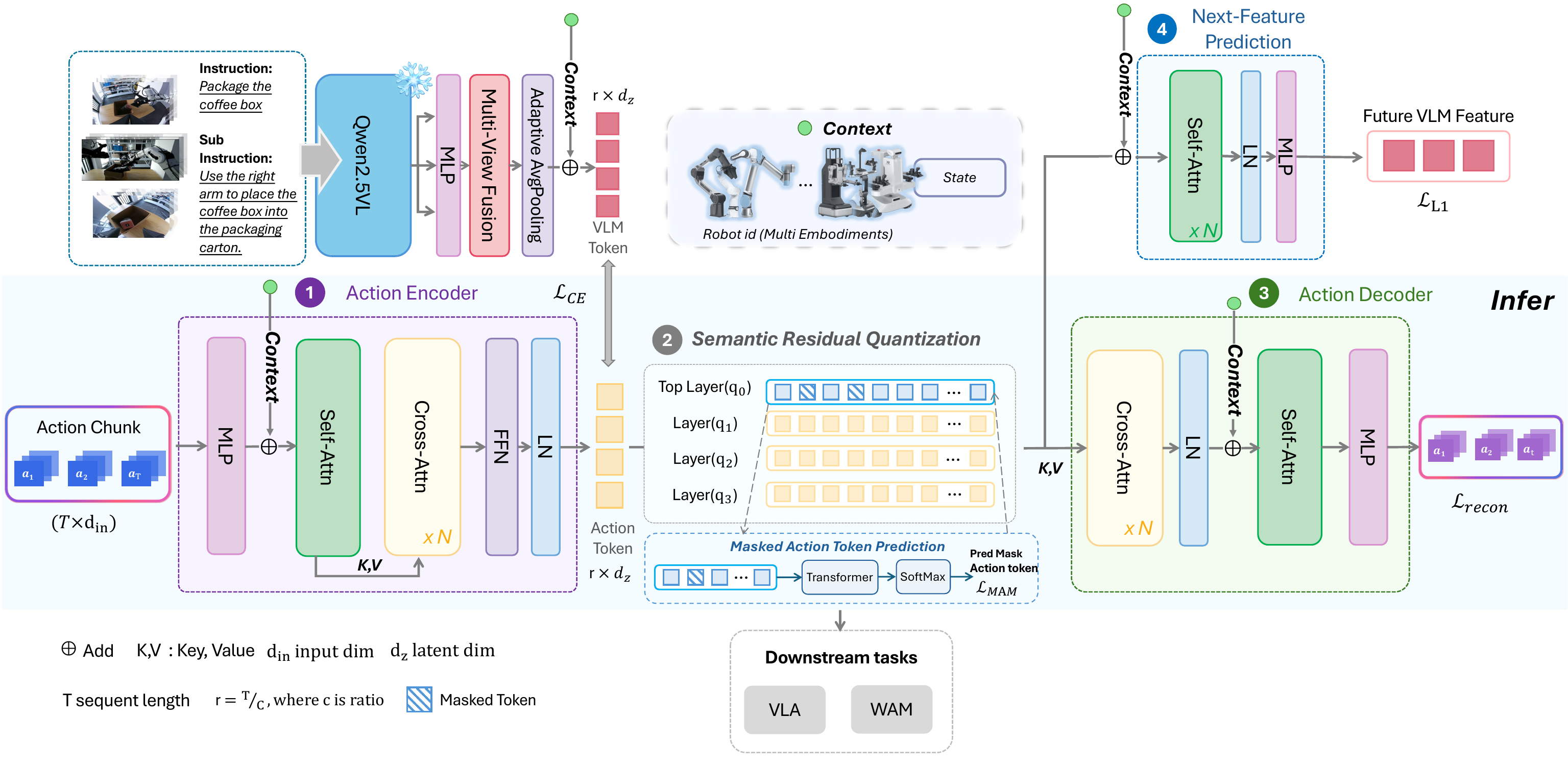}
    \vspace{-7mm}
    \caption{\small \textbf{Overview of X-Tokenizer.} Inference uses only modules \textcircled{1} Action Encoder, \textcircled{2} SRQ, and \textcircled{3} Action Decoder; pretraining additionally uses module \textcircled{4} Next-Feature Prediction and the VLM stream.}
    \label{fig:xtokenizer:overview}
    \vspace{-5mm}
\end{figure}

\subsection{Overview}
\label{sec:method:overview}

X-Tokenizer learns a discrete action space that is both reconstructive and
semantically aligned with the VLM that consumes it
(Fig.~\ref{fig:xtokenizer:overview}). Its key design principle is to
separate semantic motion intent from execution residuals across RVQ depth,
rather than treating all levels uniformly.

Given an action chunk
$\mathbf{a}_{t:t+T-1}\in\mathbb{R}^{T\times D}$~\cite{zhao2023learning},
the tokenizer defines
\begin{equation}
    \mathbf{a}_{t:t+T-1}
    \xrightarrow{E_\theta}
    h_{1:M}
    \xrightarrow{Q_\psi}
    \boldsymbol{\tau}_{1:M}
    \xrightarrow{D_\phi}
    \hat{\mathbf{a}}_{t:t+T-1},
    \label{eq:overview_pipeline}
\end{equation}
where $E_\theta$ encodes the action chunk into $M$ continuous latents,
$Q_\psi$ is the Semantic Residual Quantization (SRQ) bottleneck that maps
each latent to a multi-level discrete token, and $D_\phi$ reconstructs
executable actions. SRQ maps each latent $h_i$ to $Q$ codebook
indices $c_i^{(q)}$, yielding a discrete tuple and continuous
reconstruction
\begin{equation}
    \boldsymbol{\tau}_i=(c_i^{(1)},\ldots,c_i^{(Q)}),
    \qquad
    \tilde{\mathbf{z}}_i
    = \sum_{q=1}^{Q}\mathbf{e}^{(q)}_{c_i^{(q)}},
    \label{eq:srq_token}
\end{equation}
where $\mathbf{e}^{(q)}_j$ is the $j$-th codeword in the $q$-th codebook;
the decoder reconstructs
$\hat{\mathbf{a}}_{t:t+T-1}=D_\phi(\tilde{\mathbf{z}}_{1:M})$. Pretraining
optimizes a joint objective
\begin{equation}
    \mathcal{L}_{\mathrm{pre}}
    =
    \mathcal{L}_{\mathrm{rec}}
    +
    \lambda_{\mathrm{mam}}\mathcal{L}_{\mathrm{mam}}
    +
    \lambda_{\mathrm{align}}\mathcal{L}_{\mathrm{align}}
    +
    \lambda_{\mathrm{pred}}\mathcal{L}_{\mathrm{pred}},
    \label{eq:pretrain_loss}
\end{equation}
where $\mathcal{L}_{\mathrm{rec}}$ enforces action reconstruction,
$\mathcal{L}_{\mathrm{mam}}$ performs masked-action modeling over the
top-level discrete codes $c^{(1)}_{1:M}$, $\mathcal{L}_{\mathrm{align}}$
aligns the pre-quantization latents $h_{1:M}$ with fused VL features
derived from a frozen Qwen2.5-VL-7B extractor, and $\mathcal{L}_{\mathrm{pred}}$ acts on
the quantized latents $\tilde{\mathbf{z}}_{1:M}$ to preserve predictive
VL information. This instantiates the SRQ asymmetry: only the first RVQ
level receives discrete-level semantic supervision, while deeper levels
($q>1$) receive none.

These auxiliary heads are used only during pretraining and removed
afterward, leaving the lightweight encoder--SRQ--decoder core. This core
encodes expert trajectories offline into discrete tokens that supervise
downstream VLA co-training (\S\ref{sec:method:cotrain}); the downstream
policy does not invoke X-Tokenizer at inference time.

\subsection{Tokenizer Core Architecture: Encoder--SRQ--Decoder}
\label{sec:method:backbone}

\textbf{Encoder.}
The encoder maps a $T$-frame action chunk to $M$ continuous latents
($M\!\ll\!T$). We tokenize delta actions~\cite{feng2026demystifying,zheng2026unleashing} (per-frame offsets relative to
a proprioceptive anchor $o$ observed just before the chunk) rather than
absolute commands: absolute commands are state-dependent and vary across
embodiments, forcing a fixed-size codebook to waste capacity on
positional offsets rather than reusable motion patterns. A
Perceiver-style network~\cite{jaegle2021perceiver} downsamples the
delta-action sequence~\cite{feng2026demystifying,zheng2026unleashing} to
$M$ latent slots (by default $T\!=\!64\!\to\!M\!=\!16$) via
cross-attention from $M$ learnable queries:
$h_{1:M}=\mathrm{Enc}(x_{1:T},o,\mathbf{m})$,
where $x_{1:T}$ is the anchored delta-action chunk and $\mathbf{m}$ is a
learned embodiment token with a \emph{none} slot for CFG-style dropout,
improving robustness to unseen embodiments. Each latent slot summarizes
a coherent motion sub-segment, the semantic unit for downstream
quantization.

\textbf{Semantic Residual Quantization (SRQ).}
\label{sec:method:srq}
SRQ is the discretization bottleneck of the pipeline. We use Residual
Vector Quantization (RVQ)~\cite{lee2022autoregressive} with $Q$ stacked
levels (Eq.~\ref{eq:srq_token}), but supervise the levels asymmetrically.
In standard RVQ, every level sees the same reconstruction loss, which
tends to drive all levels toward near-uniform usage and leaves no level
with a distinct interpretable role (empirically supported by the
per-level perplexity in \S\ref{sec:exp:tokenizer:ablation}). Our asymmetric
supervision reflects the natural two-fold structure of a robot
trajectory---coarse motion intent (what the robot is doing, e.g.\
``move to the cup'') and fine geometric corrections (how exactly it
does it)---and routes each into its own RVQ layer.

\textbf{Decoder.}
The decoder reconstructs the full-length action chunk from the requantized
latent via $\hat{\mathbf{a}}_{t:t+T-1} = \mathrm{Dec}(\tilde{\mathbf{z}}_{1:M},\,o,\,\mathbf{m})$,
using a Perceiver IO-style read-out head~\cite{jaegle2022perceiver}. The
decoder is kept lightweight: most of the modelling capacity sits in the
encoder and SRQ, and the decoder only translates the discrete latent back
into executable controls. Together, the encoder, SRQ, and decoder form a
compact core that remains in the loop after all auxiliary heads are
removed at deployment, which bounds the per-call latency for offline
encoding of expert trajectories.

\subsection{Rich-Supervision Signals for Semantic Infusion}
\label{sec:method:rich_supervision}

The three pretraining heads are how X-Tokenizer acquires semantic tokens
rather than purely geometric reconstruction clusters. Each targets a
different aspect of what makes a token semantic: predictability over time
(MAM), alignment with the vision-language space (contrastive alignment),
and forward-awareness of physical consequence (next-frame VL
prediction)---syntactic regularity, semantic grounding, and predictive
physical consequence, respectively.

\textbf{Masked Action Modeling (MAM).} We apply a BERT-style
masked-prediction objective to the top-level discrete indices
$c^{(1)}_{1:M}$. A random subset $\mathcal{M}$ of positions is masked, and
a small Transformer recovers them from the surrounding context:
\begin{equation}
\mathcal{L}_{\mathrm{mam}}
\;=\; \mathbb{E}_{i \in \mathcal{M}}
\left[ -\log p_\theta\!\bigl(c^{(1)}_i \,\big|\, \tilde c^{(1)}_{1:M}\bigr) \right],
\label{eq:mam}
\end{equation}
where $\tilde c^{(1)}_{1:M}$ is the corrupted code sequence. By requiring
the top-level code stream to be predictable from its own context, MAM
turns the top-level discrete sequence into an internal action language,
while leaving the deeper layers free to specialize in reconstruction
residuals.

\textbf{Vision-Language Contrastive Alignment.}
We align the encoder's continuous latent sequence $h_{1:M}$ to fused VL
features derived from a frozen Qwen2.5-VL-7B
extractor~\cite{bai2025qwen25vltechnicalreport}. Although
$\mathcal{L}_{\mathrm{align}}$ acts on the pre-quantization $h_{1:M}$, it
reshapes the encoder feature distribution so that semantically similar
chunks cluster together; the first RVQ level's nearest-neighbor lookup
then inherits this structure, while deeper levels absorb the residual.
Let $u_{1:M}$ denote the fused multi-view VL features derived from the
frozen Qwen2.5-VL-7B extractor and temporally pooled to length $M$
(App.~\ref{appx:vl}). We apply
InfoNCE~\cite{rusak2025infonceidentifyinggaptheory,li2024decisionnce} at
two granularities within a batch of size $B$:
\begin{equation}
\mathcal{L}_{\mathrm{global}}
\;=\; -\frac{1}{B} \sum_{b=1}^B \log
\frac{\exp(\bar{h}_b \cdot \bar{u}_b / \kappa_1)}
{\sum_{b'=1}^B \exp(\bar{h}_b \cdot \bar{u}_{b'} / \kappa_1)},
\label{eq:global}
\end{equation}
\begin{equation}
\mathcal{L}_{\mathrm{local}}
\;=\; -\frac{1}{B \cdot M} \sum_{b=1}^B \sum_{i=1}^M \log
\frac{\exp(h_{b,i} \cdot u_{b,i} / \kappa_2)}
{\sum_{b'=1}^B \sum_{j=1}^M \exp(h_{b,i} \cdot u_{b',j} / \kappa_2)},
\label{eq:local}
\end{equation}
with $\bar{h}, \bar{u}$ temporally mean-pooled trajectories, $\kappa_1,
\kappa_2$ learnable temperatures, both terms symmetrized over the two
directions following CLIP~\cite{radford2021learning}, and
$\mathcal{L}_{\mathrm{align}} = \tfrac{1}{2}(\mathcal{L}_{\mathrm{global}} + \mathcal{L}_{\mathrm{local}})$.
The two granularities are complementary:
$\mathcal{L}_{\mathrm{global}}$ enforces chunk-level correspondence
between the action segment and the instruction-guided visual context via
across-chunk batch negatives, while $\mathcal{L}_{\mathrm{local}}$ binds
each slot to its time-aligned visual moment by contrasting against all
$BM\!-\!1$ other (chunk, time) pairs in the batch.

\textbf{Next-Frame VL Feature Prediction.} We attach a small auxiliary
predictor $G$ to the multi-level quantized latent
$\tilde{\mathbf{z}}_{1:M}$ that regresses the VL feature of the frame
immediately following the chunk window:
\begin{equation}
\mathcal{L}_{\mathrm{pred}}
\;=\; \bigl\|G(\tilde{\mathbf{z}}_{1:M}) - u_{+}\bigr\|_1,
\label{eq:pred}
\end{equation}
where $u_+$ is the next-frame VL feature. While MAM and the contrastive
head ground the codes in the present chunk and its current visual
context, this objective adds a forward-looking signal~\cite{assran2025v}:
the codebook is required to encode the immediate physical consequence of
the action rather than only the instantaneous geometry of the current
chunk.

\subsection{Downstream Co-training and Deployment}
\label{sec:method:cotrain}

X-Tokenizer is not a policy, but a training-time semantic scaffold for
hybrid discrete--continuous VLA policies. In this setting, a causal VLM backbone
shares hidden states $h_{\mathrm{vlm}}$ with a continuous Flow Matching action
expert. The discrete branch predicts X-Tokenizer codes autoregressively in
position-major raster order (all $Q$ levels at position $i$ before moving to
$i{+}1$), while the continuous branch regresses action trajectories:
\begin{equation}
\mathcal{L}_{\mathrm{co}}
=
-\sum_{i=1}^{M}\sum_{q=1}^{Q}
\log p_\psi\!\left(c_i^{(q)}\mid h_{\mathrm{vlm}},c_{<i}^{(1:Q)},c_i^{(<q)}\right)
+
\lambda_{\mathrm{fm}}
\mathbb{E}_{t,x_t}\!\left[
\|v_\phi(x_t,t\mid h_{\mathrm{vlm}})-u_t^\star\|_2^2
\right].
\label{eq:cotrain}
\end{equation}
The discrete loss regularizes the shared hidden states, whereas the Flow
Matching branch preserves high-fidelity continuous control. Because
X-Tokenizer codes are pretrained to align with the VLM feature space, their
prediction imposes an action-semantic supervision signal rather than a
reconstruction-only vocabulary as in FAST~\cite{pertsch2025fast}. This makes
the discrete objective improve multimodal grounding and action conditioning
for the continuous expert.

Expert trajectories are encoded offline into multi-level tokens
$\mathbf{c}=\{c^{(1)},\ldots,c^{(Q)}\}_{1:M}$. The autoregressive branch
predicts all $Q$ levels, where $c^{(1)}$ carries semantic supervision from
pretraining (\S\ref{sec:method:rich_supervision}) and deeper levels refine
action fidelity. At inference, both the autoregressive head and X-Tokenizer are
disabled, so the policy runs as a single-forward continuous flow regressor with
no discrete-token overhead.
Additional architecture details, reconstruction losses, and wall-clock latency
measurements are provided in App.~\ref{appx:xtokenizer}.

\section{Experiments}
\label{sec:exp}

We pretrain X-Tokenizer on about 2.4M trajectories and about 2.0B
action frames spanning $17$ arm families, assembled from X Square Robot-internal
data together with public academic and third-party robotic-manipulation
datasets (full corpus, embodiment registry, and baseline tokenizer
configurations in App.~\ref{appx:data}). We compare primarily
against FAST~\cite{pertsch2025fast}, the only publicly released
cross-embodiment tokenizer at comparable scale; RDT2~VQ%
~\cite{liu2026rdt2} and a non-learned $256$-bin per-channel uniform
quantizer serve as additional controls used in the noise-robustness
analysis of \S\ref{sec:exp:tokenizer:noise} and the
reconstruction-$\ell_1$ axis of \S\ref{sec:exp:tokenizer:ablation}.

The remainder of this section presents four studies: a multimodal
alignment analysis testing whether X-Tokenizer's discrete tokens live
in the same feature space as the consuming VLM
(\S\ref{sec:exp:alignment}); a codebook-side analysis of SRQ
specialization, semantic-head ablation, and deployment-time
robustness and latency (\S\ref{sec:exp:codebook_deploy}); a controlled
benchmark on RoboTwin~2.0 against published continuous-action
baselines (\S\ref{sec:exp:robotwin}); and a real-robot evaluation
comparing four action interfaces under matched data and training
schedule (\S\ref{sec:exp:wallx}).

\subsection{Multimodal Alignment with the VLM}
\label{sec:exp:alignment}

We test whether X-Tokenizer's latent space lives in the same multimodal
manifold as the consuming VLM along three complementary axes: a
statistical view via cosine similarities between action and VL features
(\S\ref{sec:exp:tokenizer:alignment}); a geometric view via a joint
UMAP projection~\cite{mcinnes2018umap} (\S\ref{sec:exp:tokenizer:umap}); and a functional view
that asks whether a frozen VL feature can drive the action decoder
through the SRQ codebook (\S\ref{sec:exp:tokenizer:functional}).

\subsubsection{Statistical Alignment at Two Granularities}
\label{sec:exp:tokenizer:alignment}

\begin{figure}[t]
\centering
    \begin{minipage}[c]{0.48\linewidth}
        \centering
        \includegraphics[width=\linewidth]{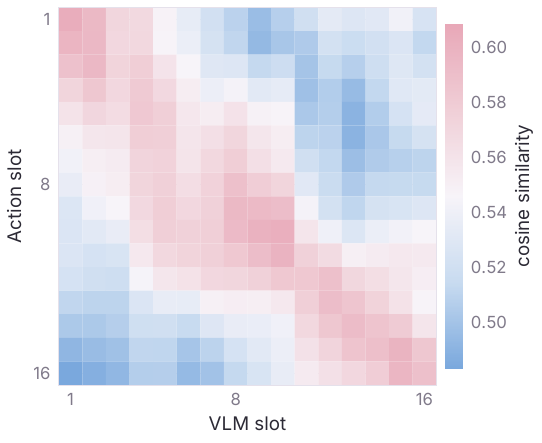}
        \subcaption{Token-level alignment.}
        \label{fig:align:perstep}
    \end{minipage}\hfill
    \begin{minipage}[c]{0.48\linewidth}
        \centering
        \includegraphics[width=\linewidth]{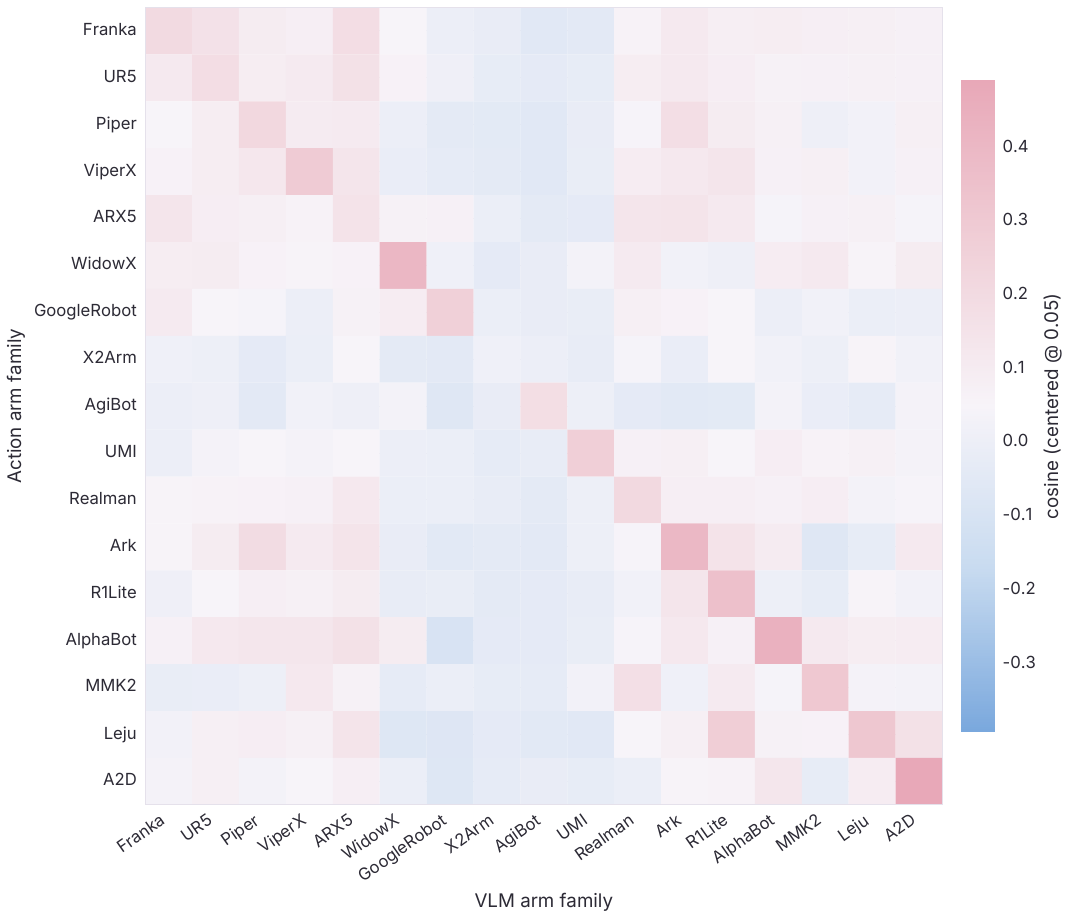}
        \subcaption{Cross arm-family alignment.}
        \label{fig:align:family}
    \end{minipage}
\caption{\small \textbf{Action--vision cosine alignment at two
granularities.} (a) per-slot cosine on length-$64$ chunks ($M{=}16$);
(b) cross-arm-family cosine matrix between sequence-pooled action and
VL features (centered at $0.05$).}
\label{fig:alignment_two_scales}
\end{figure}

For each validation chunk, we compare the encoder's pre-quantization action latent with the fused VL feature derived from a frozen Qwen2.5-VL-7B extractor for the same episode,
after pooling both to the same valid time slots and L2-normalizing the
resulting embeddings. 

The slot-level $16\!\times\!16$ heatmap of Fig.~\ref{fig:align:perstep}
shows a clear diagonal band peaking mid-chunk ($\sim\!0.60$ cosine) and
weakening at the boundaries where the VL context is partial; the
arm-family matrix of Fig.~\ref{fig:align:family} has a uniformly
positive diagonal centered at $\sim\!0.05$ above corpus mean, with
bright off-diagonal blocks between morphologically related arms.

\subsubsection{Geometric Alignment in a Shared Manifold (UMAP)}
\label{sec:exp:tokenizer:umap}

\begin{figure}[t]
    \centering
    \includegraphics[width=1\linewidth]{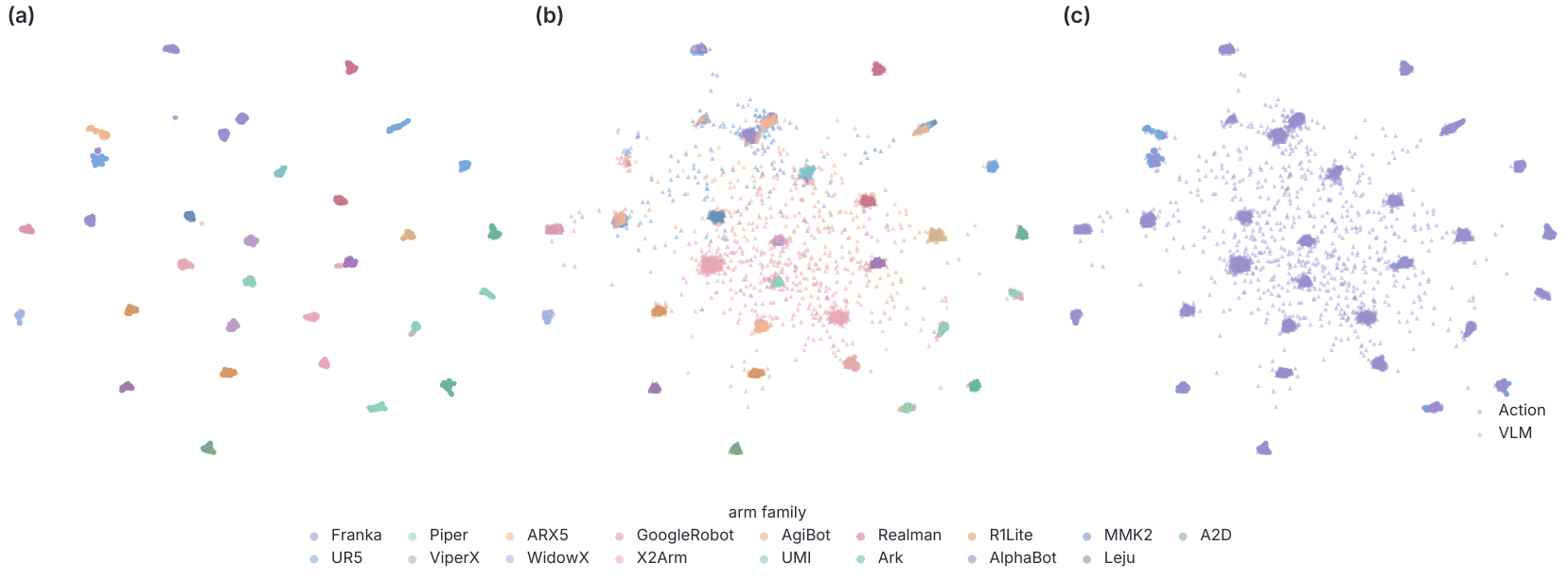}
    \caption{\small \textbf{Joint multimodal manifold via the alignment
    head.} UMAP~\cite{mcinnes2018umap} of L2-normalized sequence-mean
    alignment features. (a) action features cluster by embodiment; (b)
    VL features of the same chunks are interleaved across arms; (c)
    overlaying both modalities ($\triangle$~action, $\circ$~VLM) shows
    a single shared region.}
    \label{fig:embedding_umap}
\end{figure}

Fig.~\ref{fig:embedding_umap} visualizes the alignment space learned by
the contrastive head. Action features retain embodiment-dependent
structure, while VL features are more interleaved across arms. When
overlaid, the two modalities occupy the same broad region rather than
forming separate modality clusters. This supports the view that the
alignment head brings action and VL representations into a shared space
while preserving task- and embodiment-level variation.

\subsubsection{Functional Substitution: VLM as Action Surrogate}
\label{sec:exp:tokenizer:functional}

\begin{figure}[t]
\centering
\includegraphics[width=1\linewidth, height=6.5cm]{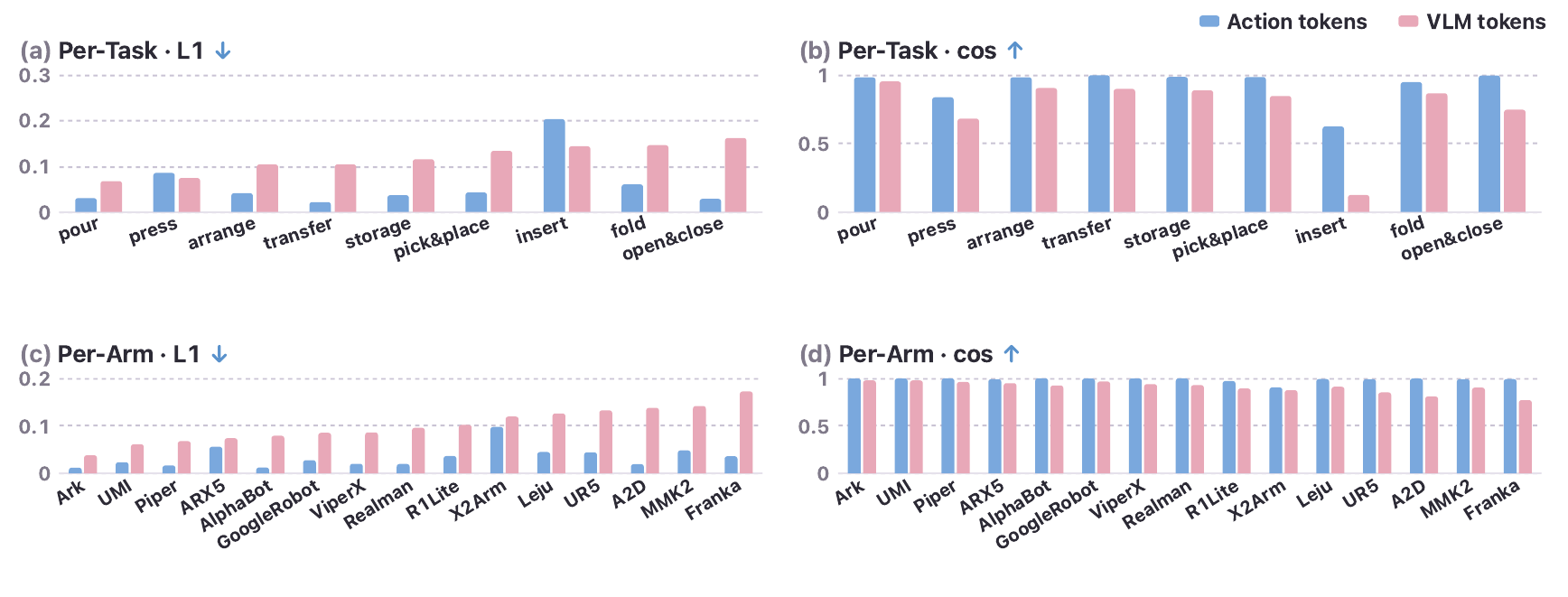}
\caption{\small \textbf{VL features as a functional surrogate.}
Reconstruction quality when the frozen SRQ + decoder is fed VL features
(purple) vs.\ action features (blue), by task (a, b) and arm family
(c, d).}
\label{fig:vlm_decode}
\end{figure}

The alignment evidence so far is statistical and geometric. We next ask
whether the aligned VL representation is usable by the action codec
itself. We route the fused VL feature $\hat{v}$ through the same
SRQ-decoder stack and compare the resulting cross-modal reconstruction
$\hat{x}^{\mathrm{vlm}}$ with the standard action-encoded
reconstruction $\hat{x}^{\mathrm{act}}$ (Fig.~\ref{fig:vlm_decode}).

The VL-driven route preserves action \emph{direction} well, reaching
per-task cosine similarity of $0.85$--$0.95$ against the
$\approx\!0.99$ action-encoded baseline, while incurring a larger
$L_1$ error. The gap is largest on fine pre-contact tasks such as
\emph{insert/plug} and \emph{press/button}, suggesting that VL features
capture the high-level motion family while the action encoder preserves
millimetre-scale execution geometry.

This functional probe shows that the learned alignment makes VL
features usable by the action codebook, rather than merely nearby in an
embedding plot. Standard action-only tokenizers such as
FAST~\cite{pertsch2025fast} or RDT-VQ~\cite{liu2026rdt2} do not provide
such a VL-to-codebook path.

\subsection{Codebook Structure and Deployment Properties}
\label{sec:exp:codebook_deploy}

Beyond multimodal alignment, we verify that the SRQ asymmetric design
behaves at the codebook level as predicted by \S\ref{sec:method:srq}
(\S\ref{sec:exp:tokenizer:codebook}), trace its dependence on each of
the three semantic heads (\S\ref{sec:exp:tokenizer:ablation}), and
characterize the deployed tokenizer's robustness to action noise and
per-call latency (\S\ref{sec:exp:tokenizer:noise}).

\subsubsection{SRQ Codebook Specialization}
\label{sec:exp:tokenizer:codebook}

\begin{figure}[t]
    \centering
    \includegraphics[width=0.85\linewidth]{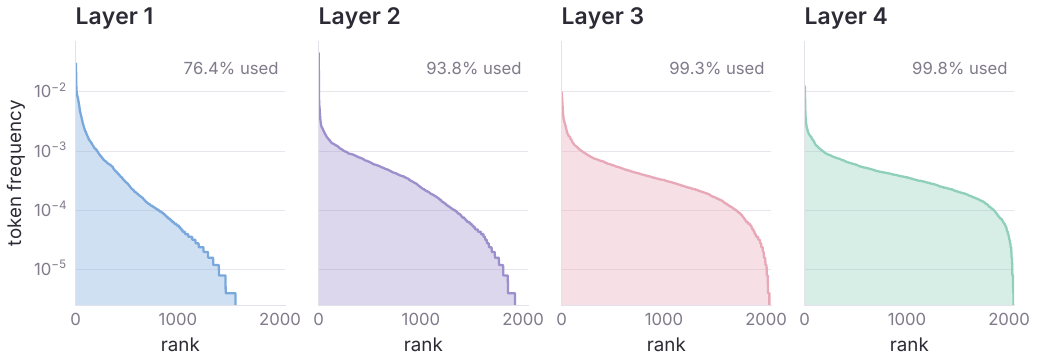}
    \caption{\small \textbf{SRQ codebook structure across the four
    residual levels.} Sorted token frequencies (log scale) on the
    validation split; per-layer usage annotated top-right of each
    panel.}
    \label{fig:codebook_usage}
\end{figure}

In a reconstruction-only RVQ, all levels are optimized to reduce
reconstruction error. Under SRQ (\S\ref{sec:method:srq}), the four
levels instead show a clear division of labor
(Fig.~\ref{fig:codebook_usage}). The MAM-regularized main code
(Layer~1) is markedly long-tailed: a small set of frequent ``motion
words'' covers the bulk of chunks, while the tail is used for less
frequent motion patterns. It still keeps $76.4\%$ of the codebook
active, with token frequencies spanning four orders of magnitude.
Layers~2--4, supervised only by reconstruction, fill the codebook much
more uniformly ($93.8\%$ / $99.3\%$ / $99.8\%$) and behave as residual
correction codes. The contrast between Layer~1 and Layers~2--4 matches the Zipf-vs-uniform pattern encouraged by SRQ, while no level shows the catastrophic collapse ($<\!10\%$ active usage) often observed in action VQ models.

\subsubsection{Ablation of Semantic Heads}
\label{sec:exp:tokenizer:ablation}

\begin{table}[H]
\centering
\footnotesize
\setlength{\tabcolsep}{3pt}
\renewcommand{\arraystretch}{1.05}
\caption{\small \textbf{Tokenizer ablation.} Reconstruction $\ell_1$
($\Delta$ vs FAST) and per-level RVQ perplexity.}
\label{tab:tokenizer_ablation}
\begin{tabular}{lcccccc}
\toprule
 & \multicolumn{2}{c}{\textbf{Recon.}} & \multicolumn{4}{c}{\textbf{RVQ PPL}} \\
\cmidrule(lr){2-3} \cmidrule(lr){4-7}
\textbf{Method} & $\bm{\ell_1}\!\downarrow$ & $\bm{\Delta}$\% & $\bm{q_0}\!\downarrow$ & $\bm{q_1}$ & $\bm{q_2}$ & $\bm{q_3}\!\uparrow$ \\
\midrule
FAST                                 & 0.01446          & --            & --            & --  & --  & --           \\
$256$-bin uniform                    & 0.00486          & $-66\%$       & --            & --  & --  & --           \\
\midrule
No aux                               & 0.00815          & $-44\%$       & 751           & 693 & 756 & 757          \\
($\text{w/o}$ Align+Pred)            & 0.00830          & $-43\%$       & 687           & 904 & 853 & 793          \\
($\text{w/o}$ MAM)                   & 0.01564          & $+8\%$        & 603           & 677 & 830 & 871          \\
\textbf{X-Tokenizer (full)}          & 0.01693          & $+17\%$       & \textbf{510}  & 700 & 828 & \textbf{916} \\
\bottomrule
\end{tabular}
\end{table}

We ablate the three pretraining-time auxiliary heads of
\S\ref{sec:method:rich_supervision}: main-code MAM, VL contrastive
alignment (\textbf{Align}), and next-frame VL feature prediction
(\textbf{Pred}). All variants share the same tokenizer architecture and
base reconstruction losses; only these auxiliary heads are toggled. We
report reconstruction $\ell_1$ together with per-level RVQ perplexity,
where lower $q_0$ perplexity indicates a more concentrated intent
codebook and higher deeper-level perplexity indicates broad residual
usage.

A successful SRQ should show increasing PPL across levels: low-PPL
$q_0$ captures recurring motion intents, while deeper levels absorb
fine residual corrections. Tab.~\ref{tab:tokenizer_ablation} shows
that neither signal alone gives the full pattern. MAM concentrates
$q_0$ but does not organize the deeper residual levels; Align+Pred
improves the deeper ordering but still lacks the strongest main-code
compression. The full model produces the intended monotone spectrum
($510 \to 700 \to 828 \to 916$), at the cost of higher reconstruction
$\ell_1$. This is the intended trade-off: the $256$-bin baseline
reconstructs well but has no learned semantic structure, whereas the
downstream experiments in \S\ref{sec:exp:robotwin}--\ref{sec:exp:wallx}
use the structure induced by SRQ.

\subsubsection{Noise Robustness and Deployment Latency}
\label{sec:exp:tokenizer:noise}

\begin{table}[H]
\centering
\footnotesize
\renewcommand{\arraystretch}{1.05}
\caption{\small \textbf{Robustness against action noise} (WER; lower is better).}
\label{tab:noise_wer}
\begin{tabular*}{0.7\linewidth}{@{\extracolsep{\fill}}rcccc@{}}
\toprule
$\sigma$ & \makecell{X-Tokenizer(ours)} & FAST & $256$-bin & \makecell{RDT2 VQ} \\
\midrule
$0.004$ & $\mathbf{0.313}$ & $0.313$ & $0.454$ & $0.325$ \\
$0.006$ & $\mathbf{0.437}$ & $0.899$ & $0.533$ & $0.439$ \\
$0.008$ & $\mathbf{0.526}$ & $1.445$ & $0.597$ & $0.549$ \\
\bottomrule
\end{tabular*}
\end{table}

We inject small Gaussian noise into physical action space before
normalization or tokenization, using the same noisy chunks for all
codecs, and report Word Error Rate (WER; Tab.~\ref{tab:noise_wer}).
X-Tokenizer obtains the lowest WER across noise levels, while FAST
degrades sharply once perturbations trigger BPE re-segmentation. Raw
WER, however, does not capture where edits occur.

\begin{figure}[H]
  \centering
  \begin{subfigure}{1\linewidth}
    \includegraphics[width=\linewidth]{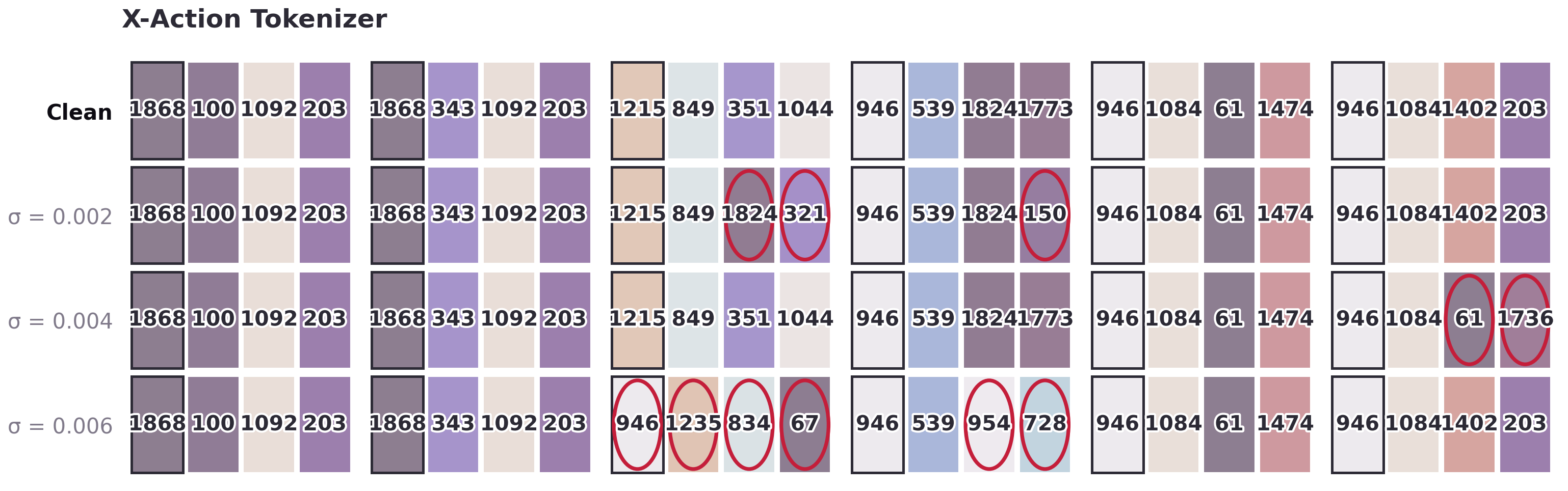}
    \caption{\small \textbf{X-Tokenizer (ours).} $24$ tokens, fixed length;
    every four cells form one residual group, with the first cell of each
    group (dark border) corresponding to the coarsest quantizer $q_0$.}
    \label{fig:codeshift_xaction}
  \end{subfigure}\\[4pt]
  \begin{subfigure}{1\linewidth}
    \includegraphics[width=\linewidth]{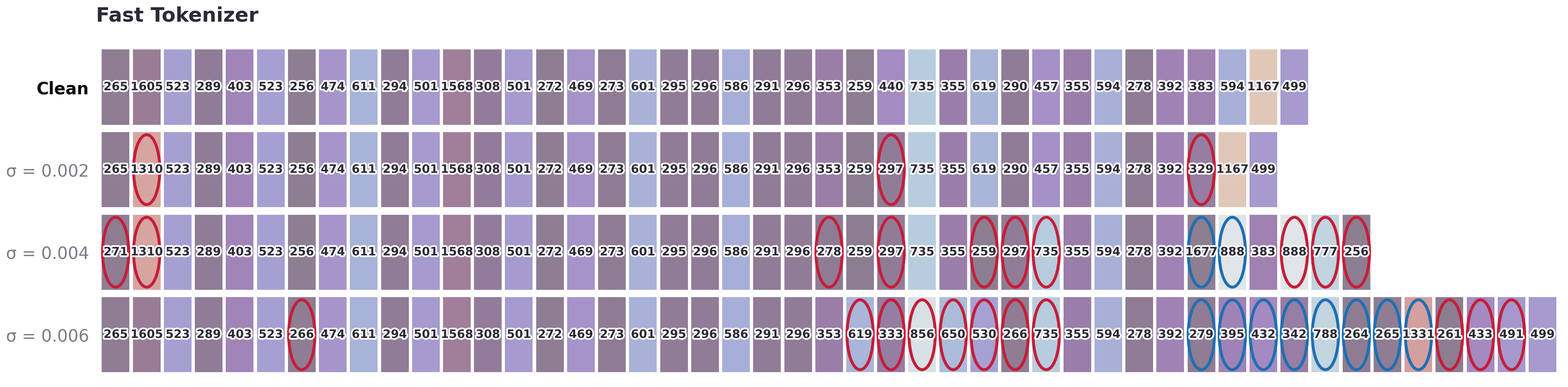}
    \caption{\small \textbf{FAST}~\cite{pertsch2025fast}. BPE-based
    variable-length codec; small noise leaves the segmentation intact,
    while larger noise re-segments the sequence and triggers many
    insertions/deletions.}
    \label{fig:codeshift_fast}
  \end{subfigure}\\[4pt]
  \begin{subfigure}{1\linewidth}
    \includegraphics[width=\linewidth]{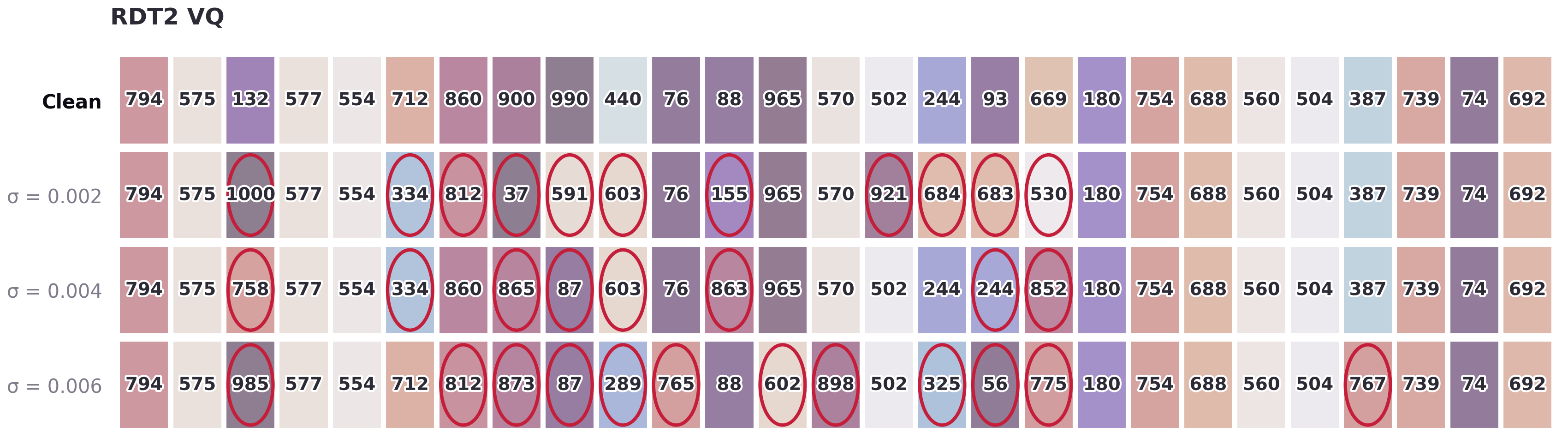}
    \caption{\small \textbf{RDT2 VQ}~\cite{liu2026rdt2}. Single-codebook
    VQ-VAE, fixed length; substitutions accumulate roughly proportionally
    to noise.}
    \label{fig:codeshift_rdt2}
  \end{subfigure}
  \caption{\small \textbf{Encoded code sequences before and after
  action-noise injection} (single chunk, all tokenizers). Top row in
  each panel is the clean reference $\mathbf{c}^{\mathrm{ref}}$;
  following rows are $\mathbf{c}^{\mathrm{hyp}}_{\sigma}$ for the three
  $\sigma$ levels. Cells are labeled with raw codebook id and
  Levenshtein-aligned edits are highlighted:
  \textcolor[HTML]{C41E3A}{red} = substitution,
  \textcolor[HTML]{1F6FB2}{blue} = insertion}
  \label{fig:codeshift}
\end{figure}

Fig.~\ref{fig:codeshift} shows where those edits land. For
X-Tokenizer, the top-level $q_0$ cells are largely stable and most
changes move into $q_{1{:}3}$, matching the SRQ hierarchy of coarse
intent plus residual execution detail. FAST instead changes sequence
length once noise perturbs its BPE segmentation, which explains the
sharp WER jump in Tab.~\ref{tab:noise_wer}. RDT2~VQ keeps a fixed
length, but its substitutions are spread across the sequence because a
single codebook does not separate intent from residual correction.
Together, the WER table and code-shift visualization show that
X-Tokenizer is robust not only by edit count, but also by preserving the
top-level intent stream under small physical perturbations. This matters
because the downstream autoregressive branch consumes the code sequence
as supervision for the shared VLM hidden states: an edit in $q_0$
changes the coarse action label seen by the backbone, whereas an edit in
$q_{1{:}3}$ mostly changes residual execution detail. SRQ therefore
turns small physical noise into lower-level corrections rather than
semantic token flips.

\begin{figure}[H]
  \centering
  \includegraphics[width=1\linewidth]{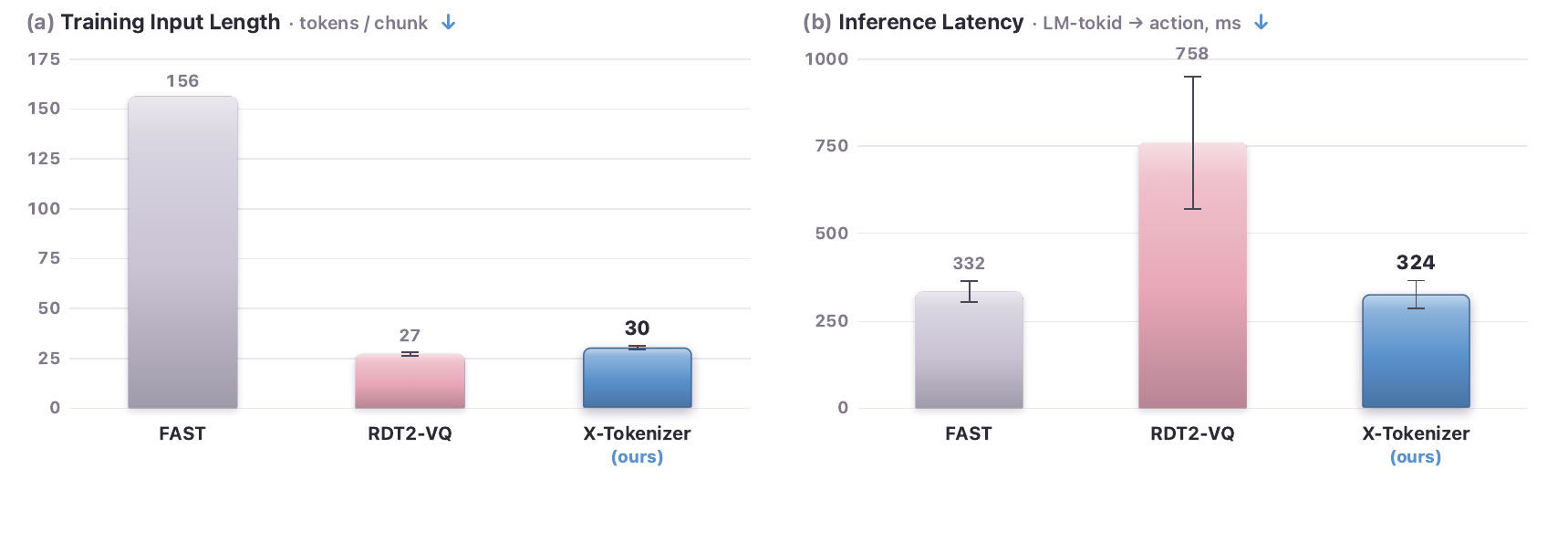}
  \vspace{-25pt}
  \caption{\small \textbf{Deployment-time tokenizer latency.} Per-chunk
  encoding latency of the three discrete codecs compared in
  Fig.~\ref{fig:codeshift} (X-Tokenizer, FAST, RDT2~VQ). For
  X-Tokenizer we measure only the deployed encoder--SRQ--decoder core
  after all pretraining-time auxiliary heads
  (\S\ref{sec:method:rich_supervision}) have been removed.}
\vspace{-15pt}
  \label{fig:tokenizer_efficiency}
\end{figure}

Beyond robustness, the deployed X-Tokenizer is also lightweight.
Fig.~\ref{fig:tokenizer_efficiency} compares per-chunk encoding
latency of the three discrete codecs on the same hardware. For
X-Tokenizer we measure only the deployed encoder--SRQ--decoder core,
after all three pretraining-time auxiliary heads have been
removed---which is what actually runs at deployment under the
asymmetric pretrain-deploy design of \S\ref{sec:method:overview}.The semantic supervision used during pretraining therefore does not
introduce extra deployment-time modules.

\subsection{RoboTwin~2.0 Benchmark}
\label{sec:exp:robotwin}

We evaluate on RoboTwin~2.0~\cite{chen2025robotwin} using the
Wall-OSS~\cite{zhai2025igniting} hybrid architecture. This benchmark
comparison uses published continuous-action baselines; the controlled
tokenizer comparison is reported in \S\ref{sec:exp:wallx}. We attach
the frozen X-Tokenizer to a released Wall-OSS checkpoint with the full
action degrees of freedom (dual arms, base, lift, head) and fine-tune
the full system for $70$k steps. The suite contains $50$ dual-arm tasks
with $50$ Clean and $500$ Randomized demonstrations per task, and each
task is evaluated with $100$ rollouts under both Easy and Hard
protocols. Full training details are provided in App.~\ref{appx:robotwin}.

\begin{figure}[H]
  \centering
  \includegraphics[width=0.85\linewidth]{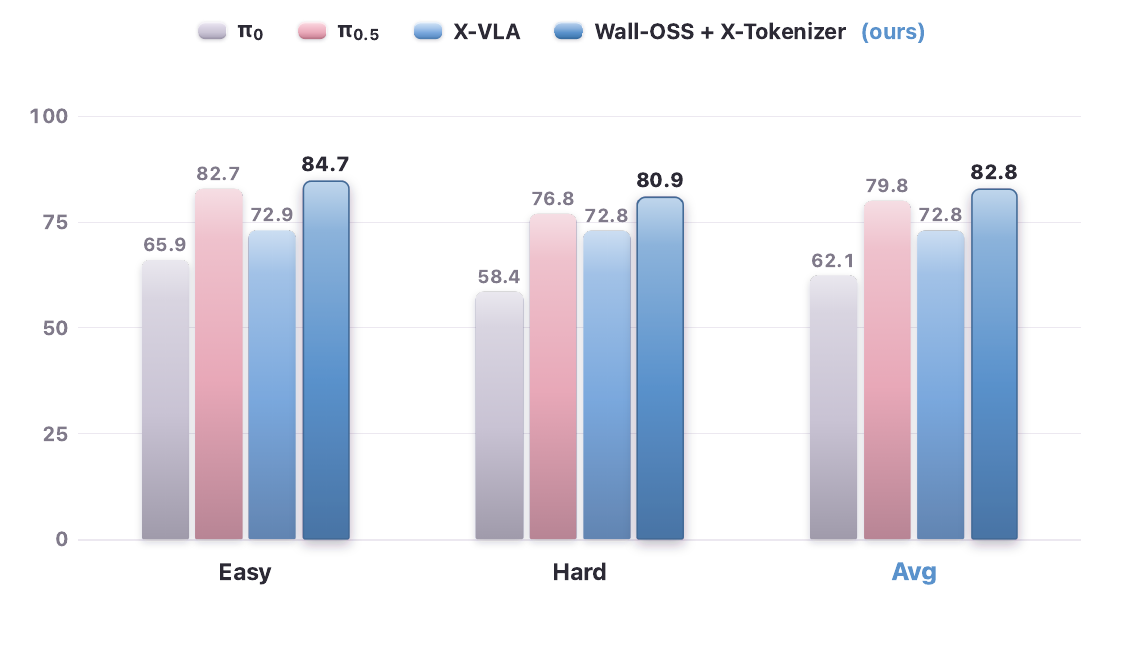}
  \caption{\textbf{RoboTwin~2.0 dual-arm} (\%).}
  \label{fig:robotwin_dualarm}
\end{figure}
\FloatBarrier

On the $50$-task dual-arm suite (Fig.~\ref{fig:robotwin_dualarm}),
Wall-OSS+X-Tokenizer achieves the best aggregate performance, improving
over the strongest published baseline $\pi_{0.5}$ in both Easy and
Hard settings. The gain is larger under Hard randomization, suggesting
that the aligned action-token interface is most useful when visual
conditions shift. Since the published methods differ in backbone,
pretraining data, and compute, we treat this as a benchmark comparison
rather than a controlled ablation.

\begin{figure}[H]
  \centering
  \includegraphics[width=0.85\linewidth]{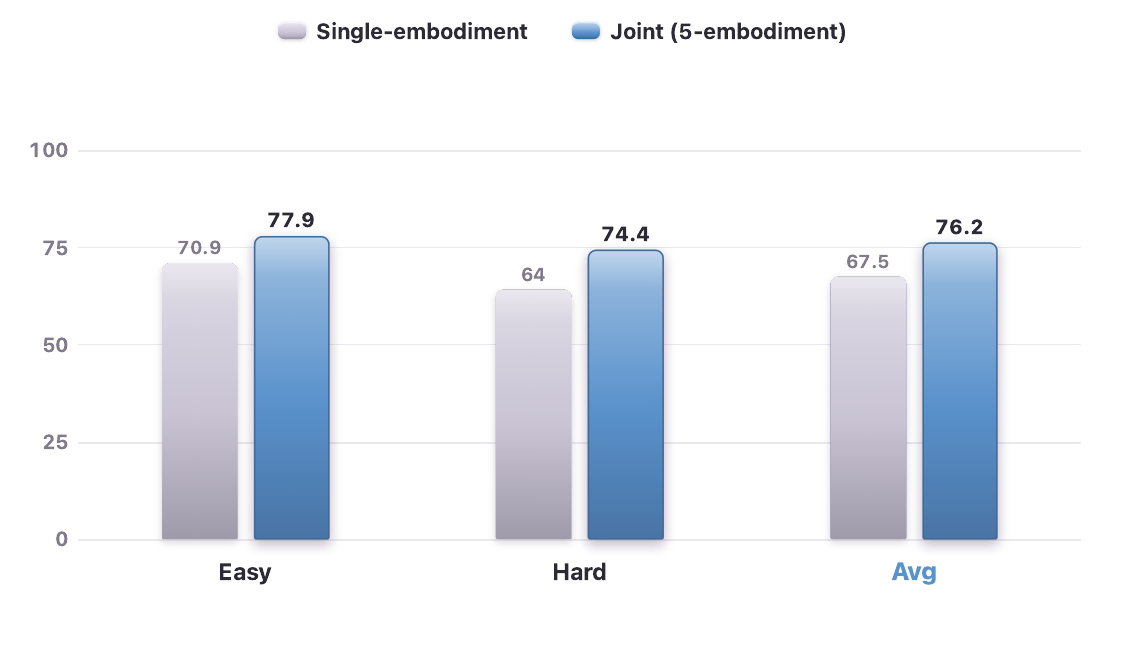}
  \vspace{-20pt}
  \caption{\textbf{Cross-embodiment (70k)} (\%).}
  \label{fig:robotwin_crossembod}
  \vspace{-20pt}
\end{figure}

To probe cross-embodiment transfer (Fig.~\ref{fig:robotwin_crossembod}),
we train on five single-arm embodiments and compare separate
single-embodiment models with one joint model trained on the union, all
for $70$k gradient steps. This controls the per-model training schedule:
the single-embodiment setting trains five separate models and therefore
uses more total compute, while each model sees only its own embodiment's
data. Joint training improves performance from $70.9\!\to\!77.9$ on
Easy and $64.0\!\to\!74.4$ on Hard, with the larger gain under harder
scene randomization.

This trend is consistent with the arm-family alignment in
Fig.~\ref{fig:align:family}: a shared action-token space can reuse
motion structure across embodiments. Increased data diversity from
joint training may also contribute, so we do not isolate the tokenizer
as the sole cause of the gain.

\subsection{Real-World Evaluation}
\label{sec:exp:wallx}

\begin{figure}[t]
    \centering
    \includegraphics[width=\linewidth]{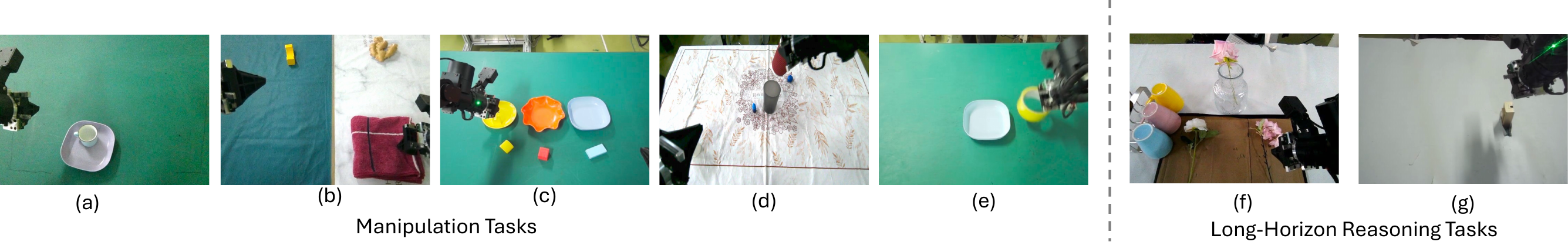}

    \includegraphics[width=\linewidth]{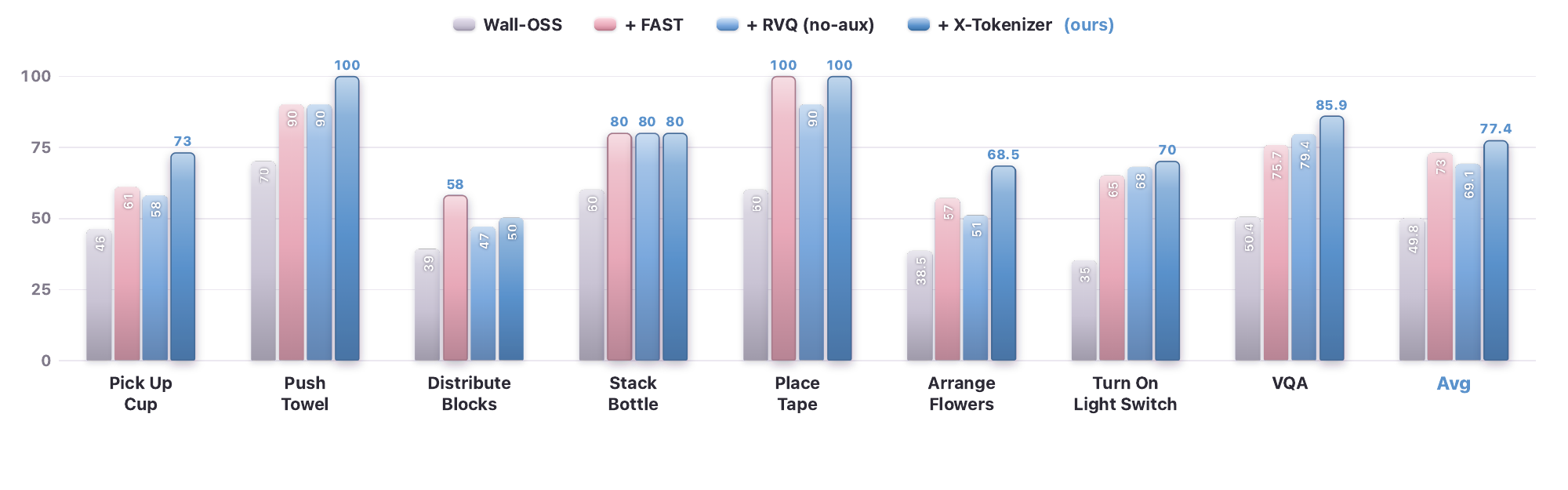}
    \caption{\footnotesize \textbf{Real-world setup and evaluation.}
    (Top) $7$ tabletop tasks: five short-horizon manipulation tasks
    (pick-up-cup, push-towel, distribute-blocks, stack-bottle,
    place-tape) and two long-horizon reasoning tasks
    (arrange-flowers, turn-on-light-switch).
    (Bottom) Per-task performance across four variants on the $7$
    tasks, plus held-out point-grounding VQA and the $7$-task average;
    the star ($\star$) marks the best (or tied-best) variant per
    column. \emph{+RVQ (no-aux)} ablates X-Tokenizer's three semantic
    heads (MAM, Align, Pred). Per-task PR is a $10$-rollout mean.}
    \label{fig:real_world_combined}
\end{figure}

We evaluate X-Tokenizer on $7$ real-world tabletop tasks
(Fig.~\ref{fig:real_world_combined}) using Wall-OSS~\cite{zhai2025igniting}
as a controlled mixed discrete-continuous VLA testbed. We compare four
action interfaces: the original Wall-OSS flow head, FAST, a
reconstruction-only $4$-level RVQ tokenizer (\emph{+RVQ no-aux}), and
our full X-Tokenizer. All variants share the same Qwen2.5-VL-3B
backbone initialization, training data, schedule, Flow Matching expert,
and evaluation protocol; only the action interface changes. The
X-Tokenizer itself is frozen from the $26$-D pretraining checkpoint.
This setting also tests cross-backbone transfer. X-Tokenizer is aligned
during pretraining to frozen Qwen2.5-VL-7B features, but is consumed
here by a Qwen2.5-VL-3B policy backbone. Each task is evaluated over
$10$ real-world rollouts using a stage-wise progress-rate (PR) rubric;
training and scoring details are in App.~\ref{appx:wallx} and
App.~\ref{appx:scoring}.
X-Tokenizer achieves the best aggregate performance:
$85.9\%$ VQA, $80.6\%$ PR over five short-horizon manipulation tasks,
$69.3\%$ PR over two long-horizon tasks, and $77.4\%$ average PR over
all seven tasks. The \emph{+RVQ no-aux} ablation is informative:
relative to FAST, it improves VQA ($75.7\!\to\!79.4$) but lowers the
$7$-task action average ($73.0\!\to\!69.1$), suggesting that
multi-level discrete structure alone helps the backbone representation
but is not sufficient for action quality. Adding MAM, Align, and Pred
raises both sides, reaching $85.9\%$ VQA and $77.4\%$ average PR.
Per-task results are consistent with this aggregate trend: X-Tokenizer
is strongest on tasks that combine manipulation with visual grounding
or multi-step instruction following, while its gains are smaller on
tasks dominated by repetitive low-level placement. This matches the
tokenizer ablation in \S\ref{sec:exp:tokenizer:ablation}: semantic
supervision improves downstream grounding and long-horizon behavior,
while paying a modest reconstruction cost.

\section{Conclusion and Future Work}
\label{sec:conclusion}

This paper argues that action tokenization for VLA pretraining should be
designed with the multimodal context in which the tokens are consumed,
rather than optimized only as action compression. Reconstruction-driven
codecs preserve trajectory geometry, but their code structure is not
explicitly aligned with the hidden states of a multimodal backbone.
X-Tokenizer addresses this with Semantic Residual Quantization and three
pretraining-time supervision heads (MAM, Align, Pred) that shape the
top-level action codes toward multimodal semantics while leaving deeper
levels to preserve execution detail. These heads are removed after
pretraining, so the deployed tokenizer remains the lightweight
encoder--SRQ--decoder core. Pretrained once on $2.4$M trajectories
($2.0$B action frames) across $17$ arm families, a single frozen
X-Tokenizer can be reused across downstream VLA settings without
tokenizer-side retraining.

The experiments support this semantic-interface view. Codebook analyses
show the intended separation between a concentrated top-level code and
broad residual levels; alignment probes show that action and fused VL
features occupy a shared representation space; and downstream results
show consistent gains in multimodal grounding and long-horizon behavior.
The reconstruction-only RVQ ablation is especially informative: hierarchy
alone improves VQA but does not recover the full action performance,
whereas adding semantic supervision improves both grounding and the
aggregate real-world task score. The transfer from a tokenizer aligned
with Qwen2.5-VL-7B features to a Qwen2.5-VL-3B policy backbone further
suggests that the learned interface is not tied to a single consuming
backbone.

Two directions extend this view. First, the current design anchors each
action chunk in end-effector space; generalizing to dexterous hands and
joint-space control would broaden the tokenizer to embodiments without a
canonical end-effector anchor. Second, the SRQ depth schedule fixes a
static reconstruction--semantics balance that could instead be adaptive
across tasks. Broadly, context-guided compression may be useful for
other discrete interfaces between foundation models and downstream
predictors, such as world-model latents.

\clearpage
\newpage
\bibliographystyle{unsrtnat}

\begin{thebibliography}{64}
\providecommand{\natexlab}[1]{#1}
\providecommand{\url}[1]{\texttt{#1}}
\expandafter\ifx\csname urlstyle\endcsname\relax
  \providecommand{\doi}[1]{doi: #1}\else
  \providecommand{\doi}{doi: \begingroup \urlstyle{rm}\Url}\fi

\bibitem[Bjorck et~al.(2025)Bjorck, Casta{\~n}eda, Cherniadev, Da, Ding, Fan, Fang, Fox, Hu, Huang, et~al.]{bjorck2025gr00t}
Johan Bjorck, Fernando Casta{\~n}eda, Nikita Cherniadev, Xingye Da, Runyu Ding, Linxi Fan, Yu~Fang, Dieter Fox, Fengyuan Hu, Spencer Huang, et~al.
\newblock Gr00t n1: An open foundation model for generalist humanoid robots.
\newblock \emph{arXiv preprint arXiv:2503.14734}, 2025.

\bibitem[Zheng et~al.(2025{\natexlab{a}})Zheng, Li, Wang, Liu, Kang, Feng, Zheng, Zou, Chen, Zeng, et~al.]{zheng2025x}
Jinliang Zheng, Jianxiong Li, Zhihao Wang, Dongxiu Liu, Xirui Kang, Yuchun Feng, Yinan Zheng, Jiayin Zou, Yilun Chen, Jia Zeng, et~al.
\newblock X-vla: Soft-prompted transformer as scalable cross-embodiment vision-language-action model.
\newblock \emph{arXiv preprint arXiv:2510.10274}, 2025{\natexlab{a}}.

\bibitem[Black et~al.(2024)Black, Brown, Driess, Esmail, Equi, Finn, Fusai, Groom, Hausman, Ichter, et~al.]{black2024pi_0}
Kevin Black, Noah Brown, Danny Driess, Adnan Esmail, Michael Equi, Chelsea Finn, Niccolo Fusai, Lachy Groom, Karol Hausman, Brian Ichter, et~al.
\newblock {$\pi_0$}: A vision-language-action flow model for general robot control.
\newblock \emph{arXiv preprint arXiv:2410.24164}, 2024.

\bibitem[Chi et~al.(2025)Chi, Xu, Feng, Cousineau, Du, Burchfiel, Tedrake, and Song]{chi2025diffusion}
Cheng Chi, Zhenjia Xu, Siyuan Feng, Eric Cousineau, Yilun Du, Benjamin Burchfiel, Russ Tedrake, and Shuran Song.
\newblock Diffusion policy: Visuomotor policy learning via action diffusion.
\newblock \emph{The International Journal of Robotics Research}, 44\penalty0 (10-11):\penalty0 1684--1704, 2025.

\bibitem[Li et~al.(2025)Li, Wu, Xi, Li, Huang, Zhang, Chen, Wang, Zhu, Liu, et~al.]{li2025controlvla}
Puhao Li, Yingying Wu, Ziheng Xi, Wanlin Li, Yuzhe Huang, Zhiyuan Zhang, Yinghan Chen, Jianan Wang, Song-Chun Zhu, Tengyu Liu, et~al.
\newblock Controlvla: Few-shot object-centric adaptation for pre-trained vision-language-action models.
\newblock \emph{arXiv preprint arXiv:2506.16211}, 2025.

\bibitem[Assran et~al.(2025)Assran, Bardes, Fan, Garrido, Howes, Muckley, Rizvi, Roberts, Sinha, Zholus, et~al.]{assran2025v}
Mido Assran, Adrien Bardes, David Fan, Quentin Garrido, Russell Howes, Matthew Muckley, Ammar Rizvi, Claire Roberts, Koustuv Sinha, Artem Zholus, et~al.
\newblock V-jepa 2: Self-supervised video models enable understanding, prediction and planning.
\newblock \emph{arXiv preprint arXiv:2506.09985}, 2025.

\bibitem[Bi et~al.(2025)Bi, Tan, Xie, Wang, Huang, Liu, Zhao, Feng, Xiang, Rong, et~al.]{bi2025motus}
Hongzhe Bi, Hengkai Tan, Shenghao Xie, Zeyuan Wang, Shuhe Huang, Haitian Liu, Ruowen Zhao, Yao Feng, Chendong Xiang, Yinze Rong, et~al.
\newblock Motus: A unified latent action world model.
\newblock \emph{arXiv preprint arXiv:2512.13030}, 2025.

\bibitem[Liu et~al.(2025{\natexlab{a}})Liu, Niu, Wang, Zheng, Zheng, Ou, Hu, Li, and Zhan]{liu2025efficient}
Dongxiu Liu, Haoyi Niu, Zhihao Wang, Jinliang Zheng, Yinan Zheng, Zhonghong Ou, Jianming Hu, Jianxiong Li, and Xianyuan Zhan.
\newblock Efficient robotic policy learning via latent space backward planning.
\newblock \emph{arXiv preprint arXiv:2505.06861}, 2025{\natexlab{a}}.

\bibitem[Maes et~al.(2026)Maes, Lidec, Scieur, LeCun, and Balestriero]{maes2026leworldmodel}
Lucas Maes, Quentin~Le Lidec, Damien Scieur, Yann LeCun, and Randall Balestriero.
\newblock Leworldmodel: Stable end-to-end joint-embedding predictive architecture from pixels.
\newblock \emph{arXiv preprint arXiv:2603.19312}, 2026.

\bibitem[Cen et~al.(2025)Cen, Yu, Yuan, Jiang, Huang, Guo, Li, Song, Luo, Wang, et~al.]{cen2025worldvla}
Jun Cen, Chaohui Yu, Hangjie Yuan, Yuming Jiang, Siteng Huang, Jiayan Guo, Xin Li, Yibing Song, Hao Luo, Fan Wang, et~al.
\newblock Worldvla: Towards autoregressive action world model.
\newblock \emph{arXiv preprint arXiv:2506.21539}, 2025.

\bibitem[Li et~al.(2026{\natexlab{a}})Li, Yao, Yang, Qu, Cheng, Yu, Lu, Von, Chen, Tang, Zhang, Ma, Li, Yang, Shu, Gao, Chen, Ye, Sun, Mon, Zhang, Li, Li, Wang, Yang, Pan, Liang, Su, Gan, Wang, and Wang]{li2026wallwmcarvingworldaction}
Shalfun Li, Victor Yao, Charles Yang, Truth Qu, Regis Cheng, Ryan Yu, Howard Lu, Newton Von, Vincent Chen, Yohann Tang, Maeve Zhang, Ellie Ma, Gody Li, Sage Yang, Lorien Shu, J.~W. Gao, Ethan Chen, Colin Ye, Yu~Sun, Elise Mon, PS~Zhang, Neo Li, Lily Li, James Wang, Ping Yang, Chris Pan, Lucy Liang, Hang Su, Roy Gan, Hao Wang, and Qian Wang.
\newblock Wall-wm: Carving world action modeling at the event joints, 2026{\natexlab{a}}.
\newblock URL \url{https://arxiv.org/abs/2606.01955}.

\bibitem[Li et~al.(2026{\natexlab{b}})Li, Zhang, Luo, Yang, Wang, Han, Yu, Gao, Xue, Zhu, et~al.]{li2026causal}
Lin Li, Qihang Zhang, Yiming Luo, Shuai Yang, Ruilin Wang, Fei Han, Mingrui Yu, Zelin Gao, Nan Xue, Xing Zhu, et~al.
\newblock Causal world modeling for robot control.
\newblock \emph{arXiv preprint arXiv:2601.21998}, 2026{\natexlab{b}}.

\bibitem[Black et~al.(2025)Black, Brown, Darpinian, Dhabalia, Driess, Esmail, Equi, Finn, Fusai, Galliker, et~al.]{black2025pi_}
Kevin Black, Noah Brown, James Darpinian, Karan Dhabalia, Danny Driess, Adnan Esmail, Michael~Robert Equi, Chelsea Finn, Niccolo Fusai, Manuel~Y Galliker, et~al.
\newblock {$\pi_{0.5}$} : a vision-language-action model with open-world generalization.
\newblock In \emph{9th Annual Conference on Robot Learning}, 2025.

\bibitem[Intelligence et~al.(2026)Intelligence, Ai, Amin, Aniceto, Balakrishna, Balke, Black, Bokinsky, Cao, Charbonnier, Choudhary, Collins, Conley, Connors, Darpinian, Dhabalia, Dhaka, DiCarlo, Driess, Equi, Esmail, Fang, Finn, Glossop, Godden, Goryachev, Groom, Habeeb, Hancock, Hausman, Hussein, Hwang, Ichter, Jacobsen, Jakubczak, Jen, Jones, Kammerer, Katz, Ke, Khadikov, Kuchi, Lamb, LeBlanc, LeCount, Levine, Li, Li-Bell, Lialin, Liang, Lim, Lu, Luo, Mano, Marwaha, Mongush, Murphy, Nair, Patterson, Pertsch, Ren, Schelske, Sharma, Shi, Shi, Smith, Springenberg, Stachowicz, Stoeckle, Tang, Tanner, Tekeste, Torne, Vedder, Vuong, Walling, Wang, Wang, Wang, Whalen, Whitmore, Williams, Xu, Yoo, Yu, Zhang, Zhang, and Zhilinsky]{intelligence2026pi07steerablegeneralistrobotic}
Physical Intelligence, Bo~Ai, Ali Amin, Raichelle Aniceto, Ashwin Balakrishna, Greg Balke, Kevin Black, George Bokinsky, Shihao Cao, Thomas Charbonnier, Vedant Choudhary, Foster Collins, Ken Conley, Grace Connors, James Darpinian, Karan Dhabalia, Maitrayee Dhaka, Jared DiCarlo, Danny Driess, Michael Equi, Adnan Esmail, Yunhao Fang, Chelsea Finn, Catherine Glossop, Thomas Godden, Ivan Goryachev, Lachlan Groom, Haroun Habeeb, Hunter Hancock, Karol Hausman, Gashon Hussein, Victor Hwang, Brian Ichter, Connor Jacobsen, Szymon Jakubczak, Rowan Jen, Tim Jones, Gregg Kammerer, Ben Katz, Liyiming Ke, Mairbek Khadikov, Chandra Kuchi, Marinda Lamb, Devin LeBlanc, Brendon LeCount, Sergey Levine, Xinyu Li, Adrian Li-Bell, Vladislav Lialin, Zhonglin Liang, Wallace Lim, Yao Lu, Enyu Luo, Vishnu Mano, Nandan Marwaha, Aikys Mongush, Liam Murphy, Suraj Nair, Tyler Patterson, Karl Pertsch, Allen~Z. Ren, Gavin Schelske, Charvi Sharma, Baifeng Shi, Lucy~Xiaoyang Shi, Laura Smith, Jost~Tobias Springenberg, Kyle Stachowicz, Will
  Stoeckle, Jiaming Tang, Jimmy Tanner, Shalom Tekeste, Marcel Torne, Kyle Vedder, Quan Vuong, Anna Walling, Haohuan Wang, Jason Wang, XuDong Wang, Chris Whalen, Samuel Whitmore, Blake Williams, Charles Xu, Sukwon Yoo, Lili Yu, Wuming Zhang, Zhuoyang Zhang, and Ury Zhilinsky.
\newblock ${\pi}_{0.7}$: a steerable generalist robotic foundation model with emergent capabilities, 2026.
\newblock URL \url{https://arxiv.org/abs/2604.15483}.

\bibitem[Yu et~al.(2026)Yu, Zhang, Liu, Liu, Kang, Li, Shi, Ma, Yang, Pan, et~al.]{yu2026wall}
Ryan Yu, Pushi Zhang, Starrick Liu, Brae Liu, Miracle Kang, Shalfun Li, Lights Shi, Ellie Ma, Ping Yang, Chris Pan, et~al.
\newblock Wall-oss-0.5 technical report.
\newblock \emph{arXiv preprint arXiv:2605.30877}, 2026.

\bibitem[Zhai et~al.(2025)Zhai, Liu, Fang, Cai, Ma, Yin, Wang, Zhou, Wang, Shi, et~al.]{zhai2025igniting}
Andy Zhai, Brae Liu, Bruno Fang, Chalse Cai, Ellie Ma, Ethan Yin, Hao Wang, Hugo Zhou, James Wang, Lights Shi, et~al.
\newblock Igniting vlms toward the embodied space.
\newblock \emph{arXiv preprint arXiv:2509.11766}, 2025.

\bibitem[Liu et~al.(2025{\natexlab{b}})Liu, Chen, An, Liu, Zhang, Gu, Li, Guo, Chen, Liu, et~al.]{liu2025hybridvla}
Jiaming Liu, Hao Chen, Pengju An, Zhuoyang Liu, Renrui Zhang, Chenyang Gu, Xiaoqi Li, Ziyu Guo, Sixiang Chen, Mengzhen Liu, et~al.
\newblock Hybridvla: Collaborative diffusion and autoregression in a unified vision-language-action model.
\newblock \emph{arXiv preprint arXiv:2503.10631}, 2025{\natexlab{b}}.

\bibitem[Zheng et~al.(2025{\natexlab{b}})Zheng, Li, Liu, Zheng, Wang, Ou, Liu, Liu, Zhang, and Zhan]{zheng2025universal}
Jinliang Zheng, Jianxiong Li, Dongxiu Liu, Yinan Zheng, Zhihao Wang, Zhonghong Ou, Yu~Liu, Jingjing Liu, Ya-Qin Zhang, and Xianyuan Zhan.
\newblock Universal actions for enhanced embodied foundation models.
\newblock In \emph{Proceedings of the Computer Vision and Pattern Recognition Conference}, pages 22508--22519, 2025{\natexlab{b}}.

\bibitem[Jiang et~al.(2025)Jiang, Yuan, Liu, Lu, Cui, Liu, Cheng, Gao, Xu, and Zhao]{jiang2025galaxea}
Tao Jiang, Tianyuan Yuan, Yicheng Liu, Chenhao Lu, Jianning Cui, Xiao Liu, Shuiqi Cheng, Jiyang Gao, Huazhe Xu, and Hang Zhao.
\newblock Galaxea open-world dataset and g0 dual-system vla model.
\newblock \emph{arXiv preprint arXiv:2509.00576}, 2025.

\bibitem[Wu et~al.(2026)Wu, Lu, Wang, Yang, Liu, Wang, Zhu, Sun, Wang, Ma, et~al.]{wu2026pragmatic}
Wei Wu, Fan Lu, Yunnan Wang, Shuai Yang, Shi Liu, Fangjing Wang, Qian Zhu, He~Sun, Yong Wang, Shuailei Ma, et~al.
\newblock A pragmatic vla foundation model.
\newblock \emph{arXiv preprint arXiv:2601.18692}, 2026.

\bibitem[Pertsch et~al.(2025)Pertsch, Stachowicz, Ichter, Driess, Nair, Vuong, Mees, Finn, and Levine]{pertsch2025fast}
Karl Pertsch, Kyle Stachowicz, Brian Ichter, Danny Driess, Suraj Nair, Quan Vuong, Oier Mees, Chelsea Finn, and Sergey Levine.
\newblock Fast: Efficient action tokenization for vision-language-action models.
\newblock \emph{arXiv preprint arXiv:2501.09747}, 2025.

\bibitem[Lee et~al.(2024)Lee, Wang, Etukuru, Kim, Shafiullah, and Pinto]{lee2024behavior}
Seungjae Lee, Yibin Wang, Haritheja Etukuru, H~Jin Kim, Nur Muhammad~Mahi Shafiullah, and Lerrel Pinto.
\newblock Behavior generation with latent actions.
\newblock \emph{arXiv preprint arXiv:2403.03181}, 2024.

\bibitem[Wang et~al.(2025)Wang, Zhu, Liu, Yang, Fang, and He]{wang2025vq}
Yating Wang, Haoyi Zhu, Mingyu Liu, Jiange Yang, Hao-Shu Fang, and Tong He.
\newblock Vq-vla: Improving vision-language-action models via scaling vector-quantized action tokenizers.
\newblock In \emph{Proceedings of the IEEE/CVF International Conference on Computer Vision}, pages 11089--11099, 2025.

\bibitem[Liu et~al.(2025{\natexlab{c}})Liu, Zhang, Dong, Ye, Yuan, Yu, Yin, Lu, Shi, Yu, et~al.]{liu2025faster}
Yicheng Liu, Shiduo Zhang, Zibin Dong, Baijun Ye, Tianyuan Yuan, Xiaopeng Yu, Linqi Yin, Chenhao Lu, Junhao Shi, Luca Jiang-Tao Yu, et~al.
\newblock Faster: Toward efficient autoregressive vision language action modeling via neural action tokenization.
\newblock \emph{arXiv preprint arXiv:2512.04952}, 2025{\natexlab{c}}.

\bibitem[Dong et~al.(2026)Dong, Liu, Zhang, Ye, Yuan, Ni, Gong, Qiu, Zhao, Li, et~al.]{dong2026actioncodec}
Zibin Dong, Yicheng Liu, Shiduo Zhang, Baijun Ye, Yifu Yuan, Fei Ni, Jingjing Gong, Xipeng Qiu, Hang Zhao, Yinchuan Li, et~al.
\newblock Actioncodec: What makes for good action tokenizers.
\newblock \emph{arXiv preprint arXiv:2602.15397}, 2026.

\bibitem[Li et~al.(2024)Li, Zheng, Zheng, Mao, Hu, Cheng, Niu, Liu, Liu, Liu, et~al.]{li2024decisionnce}
Jianxiong Li, Jinliang Zheng, Yinan Zheng, Liyuan Mao, Xiao Hu, Sijie Cheng, Haoyi Niu, Jihao Liu, Yu~Liu, Jingjing Liu, et~al.
\newblock Decisionnce: Embodied multimodal representations via implicit preference learning.
\newblock \emph{arXiv preprint arXiv:2402.18137}, 2024.

\bibitem[Radford et~al.(2021)Radford, Kim, Hallacy, Ramesh, Goh, Agarwal, Sastry, Askell, Mishkin, Clark, et~al.]{radford2021learning}
Alec Radford, Jong~Wook Kim, Chris Hallacy, Aditya Ramesh, Gabriel Goh, Sandhini Agarwal, Girish Sastry, Amanda Askell, Pamela Mishkin, Jack Clark, et~al.
\newblock Learning transferable visual models from natural language supervision.
\newblock In \emph{International conference on machine learning}, pages 8748--8763. PmLR, 2021.

\bibitem[He et~al.(2020)He, Fan, Wu, Xie, and Girshick]{he2020momentum}
Kaiming He, Haoqi Fan, Yuxin Wu, Saining Xie, and Ross Girshick.
\newblock Momentum contrast for unsupervised visual representation learning.
\newblock In \emph{Proceedings of the IEEE/CVF conference on computer vision and pattern recognition}, pages 9729--9738, 2020.

\bibitem[Lee et~al.(2022)Lee, Kim, Kim, Cho, and Han]{lee2022autoregressive}
Doyup Lee, Chiheon Kim, Saehoon Kim, Minsu Cho, and Wook-Shin Han.
\newblock Autoregressive image generation using residual quantization.
\newblock In \emph{Proceedings of the IEEE/CVF conference on computer vision and pattern recognition}, pages 11523--11532, 2022.

\bibitem[Devlin et~al.(2019{\natexlab{a}})Devlin, Chang, Lee, and Toutanova]{devlin2019bert}
Jacob Devlin, Ming-Wei Chang, Kenton Lee, and Kristina Toutanova.
\newblock Bert: Pre-training of deep bidirectional transformers for language understanding.
\newblock In \emph{Proceedings of the 2019 conference of the North American chapter of the association for computational linguistics: human language technologies, volume 1 (long and short papers)}, pages 4171--4186, 2019{\natexlab{a}}.

\bibitem[Zitkovich et~al.(2023)Zitkovich, Yu, Xu, Xu, Xiao, Xia, Wu, Wohlhart, Welker, Wahid, et~al.]{zitkovich2023rt}
Brianna Zitkovich, Tianhe Yu, Sichun Xu, Peng Xu, Ted Xiao, Fei Xia, Jialin Wu, Paul Wohlhart, Stefan Welker, Ayzaan Wahid, et~al.
\newblock Rt-2: Vision-language-action models transfer web knowledge to robotic control.
\newblock In \emph{Conference on Robot Learning}, pages 2165--2183. PMLR, 2023.

\bibitem[Kim et~al.(2024)Kim, Pertsch, Karamcheti, Xiao, Balakrishna, Nair, Rafailov, Foster, Lam, Sanketi, et~al.]{kim2024openvla}
Moo~Jin Kim, Karl Pertsch, Siddharth Karamcheti, Ted Xiao, Ashwin Balakrishna, Suraj Nair, Rafael Rafailov, Ethan Foster, Grace Lam, Pannag Sanketi, et~al.
\newblock Openvla: An open-source vision-language-action model.
\newblock \emph{arXiv preprint arXiv:2406.09246}, 2024.

\bibitem[Zhao et~al.(2023)Zhao, Kumar, Levine, and Finn]{zhao2023learning}
Tony~Z Zhao, Vikash Kumar, Sergey Levine, and Chelsea Finn.
\newblock Learning fine-grained bimanual manipulation with low-cost hardware.
\newblock \emph{arXiv preprint arXiv:2304.13705}, 2023.

\bibitem[Liu et~al.(2026{\natexlab{a}})Liu, Han, Gao, Zhao, Chen, and Du]{liu2026orderedactiontokenization}
Chaoqi Liu, Xiaoshen Han, Jiawei Gao, Yue Zhao, Haonan Chen, and Yilun Du.
\newblock Oat: Ordered action tokenization.
\newblock In \emph{Proceedings of Robotics: Science and Systems}, 2026{\natexlab{a}}.

\bibitem[Zhang et~al.(2026)Zhang, Wang, Gao, Su, Dai, Zhou, Lu, and Tang]{clap2026}
Chubin Zhang, Jianan Wang, Zifeng Gao, Yue Su, Tianru Dai, Cai Zhou, Jiwen Lu, and Yansong Tang.
\newblock Clap: Contrastive latent action pretraining for learning vision-language-action models from human videos.
\newblock \emph{arXiv preprint arXiv:2601.04061}, 2026.

\bibitem[Chen et~al.(2026)Chen, Chen, Qiu, Bai, Ge, and Ge]{unit2026}
Boyu Chen, Yi~Chen, Lu~Qiu, Jerry Bai, Yuying Ge, and Yixiao Ge.
\newblock {UniT}: Toward a unified physical language for human-to-humanoid policy learning and world modeling.
\newblock \emph{arXiv preprint arXiv:2604.19734}, 2026.

\bibitem[Feng et~al.(2026)Feng, Zheng, Wang, Liu, Li, Pang, Wang, and Zhan]{feng2026demystifying}
Yuchun Feng, Jinliang Zheng, Zhihao Wang, Dongxiu Liu, Jianxiong Li, Jiangmiao Pang, Tai Wang, and Xianyuan Zhan.
\newblock Demystifying action space design for robotic manipulation policies.
\newblock \emph{arXiv preprint arXiv:2602.23408}, 2026.

\bibitem[Zheng et~al.(2026)Zheng, Tan, Huang, Liu, Liang, Zhang, Cui, Chen, Ma, Ye, et~al.]{zheng2026unleashing}
Yinan Zheng, Tianyi Tan, Bin Huang, Enguang Liu, Ruiming Liang, Jianlin Zhang, Jianwei Cui, Guang Chen, Kun Ma, Hangjun Ye, et~al.
\newblock Unleashing the potential of diffusion models for end-to-end autonomous driving.
\newblock \emph{arXiv preprint arXiv:2602.22801}, 2026.

\bibitem[Jaegle et~al.(2021)Jaegle, Gimeno, Brock, Vinyals, Zisserman, and Carreira]{jaegle2021perceiver}
Andrew Jaegle, Felix Gimeno, Andy Brock, Oriol Vinyals, Andrew Zisserman, and Joao Carreira.
\newblock Perceiver: General perception with iterative attention.
\newblock In \emph{International Conference on Machine Learning}, pages 4651--4664. PMLR, 2021.

\bibitem[Jaegle et~al.(2022)Jaegle, Borgeaud, Alayrac, Doersch, Ionescu, Ding, Koppula, Zoran, Brock, Shelhamer, et~al.]{jaegle2022perceiver}
Andrew Jaegle, Sebastian Borgeaud, Jean-Baptiste Alayrac, Carl Doersch, Catalin Ionescu, David Ding, Skanda Koppula, Daniel Zoran, Andrew Brock, Evan Shelhamer, et~al.
\newblock Perceiver {IO}: A general architecture for structured inputs and outputs.
\newblock In \emph{International Conference on Learning Representations}, 2022.

\bibitem[Bai et~al.(2025)Bai, Chen, Liu, Wang, Ge, Song, Dang, Wang, Wang, Tang, Zhong, Zhu, Yang, Li, Wan, Wang, Ding, Fu, Xu, Ye, Zhang, Xie, Cheng, Zhang, Yang, Xu, and Lin]{bai2025qwen25vltechnicalreport}
Shuai Bai, Keqin Chen, Xuejing Liu, Jialin Wang, Wenbin Ge, Sibo Song, Kai Dang, Peng Wang, Shijie Wang, Jun Tang, Humen Zhong, Yuanzhi Zhu, Mingkun Yang, Zhaohai Li, Jianqiang Wan, Pengfei Wang, Wei Ding, Zheren Fu, Yiheng Xu, Jiabo Ye, Xi~Zhang, Tianbao Xie, Zesen Cheng, Hang Zhang, Zhibo Yang, Haiyang Xu, and Junyang Lin.
\newblock Qwen2.5-vl technical report, 2025.
\newblock URL \url{https://arxiv.org/abs/2502.13923}.

\bibitem[Rusak et~al.(2025)Rusak, Reizinger, Juhos, Bringmann, Zimmermann, and Brendel]{rusak2025infonceidentifyinggaptheory}
Evgenia Rusak, Patrik Reizinger, Attila Juhos, Oliver Bringmann, Roland~S. Zimmermann, and Wieland Brendel.
\newblock Infonce: Identifying the gap between theory and practice, 2025.
\newblock URL \url{https://arxiv.org/abs/2407.00143}.

\bibitem[Liu et~al.(2026{\natexlab{b}})Liu, Li, Ma, Wu, Tan, Ouyang, Su, and Zhu]{liu2026rdt2}
Songming Liu, Bangguo Li, Kai Ma, Lingxuan Wu, Hengkai Tan, Xiao Ouyang, Hang Su, and Jun Zhu.
\newblock Rdt2: Exploring the scaling limit of umi data towards zero-shot cross-embodiment generalization.
\newblock \emph{arXiv preprint arXiv:2602.03310}, 2026{\natexlab{b}}.

\bibitem[McInnes et~al.(2018)McInnes, Healy, and Melville]{mcinnes2018umap}
Leland McInnes, John Healy, and James Melville.
\newblock Umap: Uniform manifold approximation and projection for dimension reduction.
\newblock \emph{arXiv preprint arXiv:1802.03426}, 2018.

\bibitem[Chen et~al.(2025)Chen, Chen, Chen, Cai, Liu, Li, Liang, Lin, Ge, Gu, et~al.]{chen2025robotwin}
Tianxing Chen, Zanxin Chen, Baijun Chen, Zijian Cai, Yibin Liu, Zixuan Li, Qiwei Liang, Xianliang Lin, Yiheng Ge, Zhenyu Gu, et~al.
\newblock Robotwin 2.0: A scalable data generator and benchmark with strong domain randomization for robust bimanual robotic manipulation.
\newblock \emph{arXiv preprint arXiv:2506.18088}, 2025.

\bibitem[Gray(1984)]{gray1984vector}
Robert Gray.
\newblock Vector quantization.
\newblock \emph{IEEE Assp Magazine}, 1\penalty0 (2):\penalty0 4--29, 1984.

\bibitem[Devlin et~al.(2019{\natexlab{b}})Devlin, Chang, Lee, and Toutanova]{devlin2018bert}
Jacob Devlin, Ming-Wei Chang, Kenton Lee, and Kristina Toutanova.
\newblock {BERT}: Pre-training of deep bidirectional transformers for language understanding.
\newblock In \emph{Proceedings of the 2019 Conference of the North American Chapter of the Association for Computational Linguistics: Human Language Technologies, Volume 1 (Long and Short Papers)}, pages 4171--4186. Association for Computational Linguistics, 2019{\natexlab{b}}.

\bibitem[Shi and Hain(2021)]{shi2021contextual}
Yanpei Shi and Thomas Hain.
\newblock Contextual joint factor acoustic embeddings.
\newblock In \emph{2021 IEEE Spoken Language Technology Workshop (SLT)}, pages 750--757. IEEE, 2021.

\bibitem[Chang et~al.(2022)Chang, Zhang, Jiang, Liu, and Freeman]{chang2022maskgit}
Huiwen Chang, Han Zhang, Lu~Jiang, Ce~Liu, and William~T. Freeman.
\newblock {MaskGIT}: Masked generative image transformer.
\newblock In \emph{Proceedings of the IEEE/CVF Conference on Computer Vision and Pattern Recognition}, pages 11315--11325, 2022.

\bibitem[Hempel et~al.(2022)Hempel, Abdelrahman, and Al-Hamadi]{hempel20226d}
Thorsten Hempel, Ahmed~A Abdelrahman, and Ayoub Al-Hamadi.
\newblock 6d rotation representation for unconstrained head pose estimation.
\newblock In \emph{2022 IEEE International Conference on image processing (ICIP)}, pages 2496--2500. IEEE, 2022.

\bibitem[Mao et~al.(2019)Mao, Liu, Salzmann, and Li]{mao2019learning}
Wei Mao, Miaomiao Liu, Mathieu Salzmann, and Hongdong Li.
\newblock Learning trajectory dependencies for human motion prediction.
\newblock In \emph{Proceedings of the IEEE/CVF international conference on computer vision}, pages 9489--9497, 2019.

\bibitem[Bu et~al.(2025)Bu, Cai, Chen, Cui, Ding, Feng, He, Huang, et~al.]{bu2025agibot_iros}
Qingwen Bu, Jisong Cai, Li~Chen, Xiuqi Cui, Yan Ding, Siyuan Feng, Xindong He, Xu~Huang, et~al.
\newblock Agibot world colosseo: A large-scale manipulation platform for scalable and intelligent embodied systems.
\newblock In \emph{2025 IEEE/RSJ International Conference on Intelligent Robots and Systems (IROS)}. IEEE, 2025.

\bibitem[Team(2026)]{agibotworld2026}
AgiBot~World Team.
\newblock Agibot world 2026.
\newblock \url{https://huggingface.co/datasets/agibot-world/AgiBotWorld2026}, 2026.

\bibitem[Khazatsky et~al.(2024)Khazatsky, Pertsch, Nair, Balakrishna, Dasari, Karamcheti, Nasiriany, Srirama, Chen, Ellis, Fagan, Hejna, Itkina, Lepert, Ma, Miller, Wu, Belkhale, Dass, Ha, Jain, Lee, Lee, Memmel, Park, Radosavovic, Wang, Zhan, Black, Chi, Hatch, Lin, Lu, Mercat, Rehman, Sanketi, Sharma, Simpson, Vuong, Walke, Wulfe, Xiao, Yang, Yavary, Zhao, Agia, Baijal, Castro, Chen, Chen, Chung, Drake, Foster, Gao, Guizilini, Herrera, Heo, Hsu, Hu, Irshad, Jackson, Le, Li, Lin, Lin, Ma, Maddukuri, Mirchandani, Morton, Nguyen, O'Neill, Scalise, Seale, Son, Tian, Tran, Wang, Wu, Xie, Yang, Yin, Zhang, Bastani, Berseth, Bohg, Goldberg, Gupta, Gupta, Jayaraman, Lim, Malik, Martín-Martín, Ramamoorthy, Sadigh, Song, Wu, Yip, Zhu, Kollar, Levine, and Finn]{khazatsky2024droid}
Alexander Khazatsky, Karl Pertsch, Suraj Nair, Ashwin Balakrishna, Sudeep Dasari, Siddharth Karamcheti, Soroush Nasiriany, Mohan~Kumar Srirama, Lawrence~Yunliang Chen, Kirsty Ellis, Peter~David Fagan, Joey Hejna, Masha Itkina, Marion Lepert, Yecheng~Jason Ma, Patrick~Tree Miller, Jimmy Wu, Suneel Belkhale, Shivin Dass, Huy Ha, Arhan Jain, Abraham Lee, Youngwoon Lee, Marius Memmel, Sungjae Park, Ilija Radosavovic, Kaiyuan Wang, Albert Zhan, Kevin Black, Cheng Chi, Kyle~Beltran Hatch, Shan Lin, Jingpei Lu, Jean Mercat, Abdul Rehman, Pannag~R Sanketi, Archit Sharma, Cody Simpson, Quan Vuong, Homer~Rich Walke, Blake Wulfe, Ted Xiao, Jonathan~Heewon Yang, Arefeh Yavary, Tony~Z. Zhao, Christopher Agia, Rohan Baijal, Mateo~Guaman Castro, Daphne Chen, Qiuyu Chen, Trinity Chung, Jaimyn Drake, Ethan~Paul Foster, Jensen Gao, Vitor Guizilini, David~Antonio Herrera, Minho Heo, Kyle Hsu, Jiaheng Hu, Muhammad~Zubair Irshad, Donovon Jackson, Charlotte Le, Yunshuang Li, Kevin Lin, Roy Lin, Zehan Ma, Abhiram Maddukuri, Suvir
  Mirchandani, Daniel Morton, Tony Nguyen, Abigail O'Neill, Rosario Scalise, Derick Seale, Victor Son, Stephen Tian, Emi Tran, Andrew~E. Wang, Yilin Wu, Annie Xie, Jingyun Yang, Patrick Yin, Yunchu Zhang, Osbert Bastani, Glen Berseth, Jeannette Bohg, Ken Goldberg, Abhinav Gupta, Abhishek Gupta, Dinesh Jayaraman, Joseph~J Lim, Jitendra Malik, Roberto Martín-Martín, Subramanian Ramamoorthy, Dorsa Sadigh, Shuran Song, Jiajun Wu, Michael~C. Yip, Yuke Zhu, Thomas Kollar, Sergey Levine, and Chelsea Finn.
\newblock Droid: A large-scale in-the-wild robot manipulation dataset.
\newblock 2024.

\bibitem[Wu et~al.(2024)Wu, Hou, Liu, Che, Ju, Yang, Li, Zhao, Xu, Yang, et~al.]{wu2024robomind}
Kun Wu, Chengkai Hou, Jiaming Liu, Zhengping Che, Xiaozhu Ju, Zhuqin Yang, Meng Li, Yinuo Zhao, Zhiyuan Xu, Guang Yang, et~al.
\newblock Robomind: Benchmark on multi-embodiment intelligence normative data for robot manipulation.
\newblock \emph{arXiv preprint arXiv:2412.13877}, 2024.

\bibitem[Hou et~al.(2025)Hou, Wu, Liu, Che, Wu, Liao, Li, He, Feng, Jin, et~al.]{hou2025robomind}
Chengkai Hou, Kun Wu, Jiaming Liu, Zhengping Che, Di~Wu, Fei Liao, Guangrun Li, Jingyang He, Qiuxuan Feng, Zhao Jin, et~al.
\newblock Robomind 2.0: A multimodal, bimanual mobile manipulation dataset for generalizable embodied intelligence.
\newblock \emph{arXiv preprint arXiv:2512.24653}, 2025.

\bibitem[Wu et~al.(2025)Wu, Liu, Xie, Wang, Li, Yang, Li, Zhu, Wu, Liu, et~al.]{wu2025robocoin}
Shihan Wu, Xuecheng Liu, Shaoxuan Xie, Pengwei Wang, Xinghang Li, Bowen Yang, Zhe Li, Kai Zhu, Hongyu Wu, Yiheng Liu, et~al.
\newblock Robocoin: An open-sourced bimanual robotic data collection for integrated manipulation.
\newblock \emph{arXiv preprint arXiv:2511.17441}, 2025.

\bibitem[{RoboChallenge.ai}(2025)]{robochallenge2025}
{RoboChallenge.ai}.
\newblock {RoboChallenge Table30 v2 Dataset}.
\newblock \url{https://huggingface.co/datasets/RoboChallenge/Table30v2}, 2025.
\newblock Accessed: 2026-05-07.

\bibitem[{GenRobot AI}(2025)]{realomni2025}
{GenRobot AI}.
\newblock {10Kh RealOmni-Open DataSet}.
\newblock \url{https://www.genrobot.ai/data/open-dataset}, 2025.
\newblock Accessed: 2026-05-07.

\bibitem[Walke et~al.(2023)Walke, Black, Zhao, Vuong, Zheng, Hansen-Estruch, He, Myers, Kim, Du, et~al.]{walke2023bridgedata}
Homer~Rich Walke, Kevin Black, Tony~Z Zhao, Quan Vuong, Chongyi Zheng, Philippe Hansen-Estruch, Andre~Wang He, Vivek Myers, Moo~Jin Kim, Max Du, et~al.
\newblock Bridgedata v2: A dataset for robot learning at scale.
\newblock In \emph{Conference on Robot Learning}, pages 1723--1736. PMLR, 2023.

\bibitem[Brohan et~al.(2023)Brohan, Brown, Carbajal, Chebotar, Dabis, Finn, Gopalakrishnan, Hausman, Herzog, Hsu, et~al.]{brohan2022rt1}
Anthony Brohan, Noah Brown, Justice Carbajal, Yevgen Chebotar, Joseph Dabis, Chelsea Finn, Keerthana Gopalakrishnan, Karol Hausman, Alex Herzog, Jasmine Hsu, et~al.
\newblock {RT-1}: Robotics transformer for real-world control at scale.
\newblock In \emph{Robotics: Science and Systems (RSS)}, 2023.

\bibitem[Jang et~al.(2022)Jang, Irpan, Khansari, Kappler, Ebert, Lynch, Levine, and Finn]{jang2022bc}
Eric Jang, Alex Irpan, Mohi Khansari, Daniel Kappler, Frederik Ebert, Corey Lynch, Sergey Levine, and Chelsea Finn.
\newblock Bc-z: Zero-shot task generalization with robotic imitation learning.
\newblock In \emph{conference on Robot Learning}, pages 991--1002. PMLR, 2022.

\bibitem[Heo et~al.(2025)Heo, Lee, Lee, and Lim]{heo2025furniturebench}
Minho Heo, Youngwoon Lee, Doohyun Lee, and Joseph~J Lim.
\newblock Furniturebench: Reproducible real-world benchmark for long-horizon complex manipulation.
\newblock \emph{The International Journal of Robotics Research}, 44\penalty0 (10-11):\penalty0 1863--1891, 2025.

\bibitem[O’Neill et~al.(2024)O’Neill, Rehman, Maddukuri, Gupta, Padalkar, Lee, Pooley, Gupta, Mandlekar, Jain, et~al.]{o2024open}
Abby O’Neill, Abdul Rehman, Abhiram Maddukuri, Abhishek Gupta, Abhishek Padalkar, Abraham Lee, Acorn Pooley, Agrim Gupta, Ajay Mandlekar, Ajinkya Jain, et~al.
\newblock Open x-embodiment: Robotic learning datasets and rt-x models: Open x-embodiment collaboration 0.
\newblock In \emph{2024 IEEE International Conference on Robotics and Automation (ICRA)}, pages 6892--6903. IEEE, 2024.

\end{thebibliography}

\clearpage
\appendix
\FloatBarrier
\section{X-Tokenizer Implementation Details}
\label{appx:xtokenizer}

This appendix documents the X-Tokenizer architecture, the vision-language
feature extraction pipeline, and the implementation of every loss term
used in pretraining. The three semantic supervisions of the main text
(MAM, VL contrastive alignment, next-frame VL prediction) are documented
in App.~\ref{appx:mam}--\ref{appx:pred}; the remaining reconstruction-side
losses (rotation geodesic, frequency-domain, temporal smoothness) are
auxiliary stability regularizers and are documented in
App.~\ref{appx:geo}--\ref{appx:dctsmooth}.

\subsection{Delta-Action Backbone}
\label{appx:backbone}

\textbf{Delta action and per-channel choices.} We tokenize delta
actions---per-frame motion offsets relative to the proprioceptive anchor
$o$ observed just before the chunk---rather than absolute commands; the
$D{=}26$ channel layout and per-channel choice of $\Delta$ are listed in
Tab.~\ref{tab:appx:layout}. The choice of $\Delta$ follows physical
meaning: end-effector position and 6D rotation are configuration-space
quantities for which the chunk anchor provides a natural reference;
gripper opening, lift, and head pitch/yaw are state-like signals whose
absolute value is what downstream control consumes; base velocity is
already a temporal derivative, so an additional $\Delta$ would be
ill-posed.

\begin{table}[H]
\small
\centering
\caption{The $D{=}26$ channel layout and per-channel choice of $\Delta$.}
\label{tab:appx:layout}
\renewcommand{\arraystretch}{1.05}
\begin{tabular}{@{}llcl@{}}
\toprule
\textbf{Channel group} & \textbf{Dim} & \textbf{$\Delta$ type} & \textbf{Reason} \\
\midrule
Left/right end-effector position & $3+3=6$  & Euclidean subtraction       & Cartesian position, anchorable to $o$        \\
Left/right 6D rotation           & $6+6=12$ & $\mathrm{SO}(3)$ composition & Orientation lives on a manifold              \\
Left/right gripper               & $1+1=2$  & Identity                    & State-like (open/close)                       \\
Base velocity                    & $3$      & Identity                    & Already a temporal derivative                 \\
Lift / height                    & $1$      & Identity                    & State-like                                    \\
Head pitch + yaw                 & $2$      & Identity                    & State-like (yaw/pitch absolute angles)        \\
\midrule
\textbf{Total}                   & $\mathbf{26}$ & & \\
\bottomrule
\end{tabular}
\end{table}

\textbf{Per-channel normalization.} Per-channel MinMax statistics from a
dataset-level table---using the $0.1\%$/$99.9\%$ quantile range to define a
robust min/max for each channel---normalize each channel to $[-1,1]$. The
same table is later used to recover physical deltas for the rotation
geodesic loss (App.~\ref{appx:geo}).

\textbf{Encoder.} The encoder $\mathrm{Enc}$ ingests the delta-action
chunk $x_{1:T}$ together with the proprioceptive anchor $o$ and the
embodiment token $\mathbf{m}$, and outputs $M$ continuous latents
$h_{1:M}$ with hidden dim $H{=}1024$ at compression ratio
$r{=}4$ (so $M{=}T/r$). Concretely:
\begin{enumerate}[topsep=2pt,itemsep=1pt,partopsep=0pt,parsep=1pt,leftmargin=1.5em]
\item \textbf{Input projection.} A linear projection followed by LayerNorm, GELU and dropout maps $x_{1:T}$ from $\mathbb{R}^{D}$ to $\mathbb{R}^{H}$.
\item \textbf{Embodiment conditioning.} An encoder-side embedding vector $\mathbf{m}\!\in\!\mathbb{R}^{H}$, looked up from a learnable registry of $1024$ slots (one of which is a special learnable ``none'' slot used under CFG-style dropout), is added broadcast over time.
\item \textbf{RoPE positional encoding.} Rotary position embeddings on the time dimension with base $10^{4}$.
\item \textbf{Self-attention stack.} A $12$-layer Transformer encoder ($8$ heads, GELU FFN of width $4H$, dropout $0.1$) processes the projected sequence with the chunk's padding mask.
\item \textbf{Optional state cross-attention.} When $o$ is provided (i.e., not CFG-dropped), a single cross-attention block uses the linearly projected $o$ as key and value while the time series acts as query, followed by residual + LayerNorm.
\item \textbf{Latent query cross-attention.} $M_{\max}{=}16$ learnable latent queries $\mathbf{q}_{1:M}$ are equipped with their own RoPE encoding, expanded across the batch, and cross-attend to the encoded sequence to extract a length-$M$ summary.
\item \textbf{Position-wise FFN.} A final FFN with residual + LayerNorm.
\end{enumerate}

\textbf{Decoder.} The decoder $\mathrm{Dec}$ ingests the quantized latent
$\tilde{\mathbf{z}}_{1:M}$ together with $o$ and $\mathbf{m}$, and outputs
the reconstructed delta-action sequence $\hat{x}_{1:T}$; the final action
chunk $\hat{\mathbf{a}}_{t:t+T-1}$ is recovered by re-anchoring
$\hat{x}_{1:T}$ to $o$. We allocate $T_{\max}{=}64$ learnable position
queries $\mathbf{p}_{1:T}$ with RoPE; a decoder-side embodiment embedding
(independent from the encoder's) is added to the queries. The queries
first cross-attend to RoPE-positioned latents, then optionally to the
projected state $o$, and the result passes through a $4$-layer
self-attention Transformer ($8$ heads, dropout $0.1$). A linear output
head produces $\hat{x}_{1:T}$, after which the DoF mask zeroes out
invalid channels.

\textbf{CFG-style dropout.} At training time each conditioning signal is
independently corrupted: the observation $o$ is zeroed with probability
$p_o{=}0.2$; the embodiment id is dropped with probability $p_e{=}0.2$;
with an additional probability $p_n{=}0.1$ it is replaced by the
learnable ``none'' slot of the embodiment registry. The deployed
tokenizer is therefore robust to missing state and to embodiments outside
the registry.

\subsection{Semantic Residual Quantization}
\label{appx:srq}

SRQ is the discretization bottleneck of the tokenizer: a residual vector
quantizer that maps each continuous latent $h_i$ to a tuple of
$Q$ codebook indices, with a top-vs-deeper asymmetry in supervision (top
level exposed to semantic losses, deeper levels to reconstruction only).
The quantization itself is standard RVQ: each level $q$ owns a codebook
$\mathcal{C}^{(q)}{=}\{\mathbf{e}^{(q)}_1,\ldots,\mathbf{e}^{(q)}_V\}$,
and the quantized latent is the sum of selected codewords
$\tilde{\mathbf{z}}_i = \sum_{q=1}^{Q}\mathbf{e}^{(q)}_{c_i^{(q)}}$.

We use $Q{=}4$ levels and $V{=}2048$ codewords per level. Codebooks are
EMA-updated~\cite{gray1984vector} (decay $0.8$, dead-code reset threshold
$2$), initialized by $k$-means with $100$ iterations on the first
warmed-up batch, and use Euclidean (not cosine) similarity for
nearest-neighbor lookup. Beyond the standard commitment loss (weight
$\lambda_{\mathrm{vq}}$ in App.~\ref{appx:training}), deeper levels
receive no auxiliary supervision.

\subsection{Vision-Language Feature Extraction}
\label{appx:vl}

The VL features $u_{1:M}$ used as targets by the contrastive alignment
loss $\mathcal{L}_{\mathrm{align}}$, together with the next-frame target
$u_+$ used by the next-frame prediction loss $\mathcal{L}_{\mathrm{pred}}$,
are pre-extracted offline once and stored on disk; the deployed tokenizer
never invokes Qwen2.5-VL. The extractor is built on
Qwen2.5-VL-7B~\cite{bai2025qwen25vltechnicalreport} run at bf16 with
FlashAttention-2.

\textbf{Two-level language conditioning.} Each trajectory is annotated
with a global instruction $l^{\mathrm{inst}}$ and a list of segments
$\{(t^{(s)}_{\mathrm{start}},\,t^{(s)}_{\mathrm{end}},\,l^{\mathrm{sub},(s)})\}$
that partition the trajectory into local subtasks. The instruction fixes
the global task semantics of the entire trajectory while the segment-%
level subtask disambiguates the local intent of the current motion. For
each segment the prompt fed to the VLM is
\begin{quote}
\texttt{Task: }$l^{\mathrm{inst}}$\\
\texttt{Current step: }$l^{\mathrm{sub},(s)}$
\end{quote}
which together with the corresponding multi-view frames forms a single
forward pass. Each per-frame feature thus reflects what the robot sees
and what it is currently trying to do under the global goal. This is
also what makes the alignment target ``multimodal'' rather than purely
visual.

\textbf{Per-frame, per-view feature.} Frames are sampled at a temporal
stride of $r_v{=}4$, the image processor is configured with
$\texttt{max\_pixels}{=}200{,}704$ ($\approx448\!\times\!448$, $\sim\!256$
visual tokens per frame), and segments longer than
$\texttt{max\_segment\_vl\_frames}{=}128$ are split into sub-windows. For
each sub-window of $N$ frames and view $k\!\in\!\{1,2,3\}$ (face, left
wrist, right wrist), we read the hidden states of layer $-3$---which we
found to give cleaner spatial structure than the very last layer for
image tokens---and split them into per-image visual tokens (using the
model's spatial-merge factor) and the remaining text tokens. We average
the visual tokens of each frame to a single visual feature
$\bar v^{(k)}_t\!\in\!\mathbb{R}^{H_{\mathrm{vl}}}$, average all text
tokens to a single sub-window-level feature
$\bar w^{(s)}\!\in\!\mathbb{R}^{H_{\mathrm{vl}}}$ shared across the
sub-window, and define the per-frame VL feature as
\[
u^{(k)}_t \;=\; \tfrac{1}{2}\bigl(\bar v^{(k)}_t + \bar w^{(s)}\bigr),
\qquad H_{\mathrm{vl}}{=}3584.
\]
The result of one episode is a $T_v\!\times\!H_{\mathrm{vl}}$ tensor per
view, written to disk and indexed alongside the action chunks.

\textbf{Multi-view fusion at training time.} At training time each of the
three views is projected from $H_{\mathrm{vl}}$ to $H$ by a shared linear
layer, producing $\{u^{(k)}_t\}\!\in\!\mathbb{R}^{H}$. The three projected
streams are combined into the single fused stream $u_t$ used as alignment
target by a learned per-sample per-view weighting: a global vector
$w\!\in\!\mathbb{R}^{3}$ of unnormalized weights is set to $-\infty$ on
each sample's missing views (so that different samples within the same
batch may have different active view masks), then softmaxed; the fused
feature is $u_t\!=\!\sum_k \alpha_k u^{(k)}_t$ with
$\alpha = \mathrm{softmax}(w)$ computed per sample.

\subsection{Masked Action Modeling (MAM) Head}
\label{appx:mam}

MAM is the discrete-level semantic supervision on the top RVQ codebook:
random positions in the top-level code stream $c^{(1)}_{1:M}$ are masked,
and a small Transformer predicts them from the corrupted context $\tilde
c^{(1)}_{1:M}$ via cross-entropy.

\textbf{Head architecture.} The main code stream
$c^{(1)}_{1:M}\!\in\!\{1,\ldots,V\}^{M}$ is fed to a $2$-layer Transformer
encoder ($4$ heads, dropout $0.1$) with code-token and absolute-position
embeddings (max length $M_{\max}{=}16$). For each chunk we sample the
mask set $\mathcal{M}\!\subseteq\!\{1,\ldots,M\}$ at probability $0.15$
over valid (non-padding) positions; if a sample's mask comes out empty,
one valid position is forced to be selected so that the masked-position
cross-entropy is well defined. Following BERT-style corruption of
discrete tokens~\cite{devlin2018bert,shi2021contextual,chang2022maskgit},
the masked positions are corrupted with $80\%$ replacement by a learnable
\texttt{[MASK]} embedding, $10\%$ replacement by a uniform random codebook
entry, and $10\%$ unchanged. The Transformer's classifier projects to a
logit over the $V$ codes; this defines the predictive distribution
$p_\theta(\cdot\mid\tilde c^{(1)}_{1:M})$.

\textbf{Warm-up.} The codebooks are still moving in the early epochs, and
we find that the MAM objective is unstable when applied immediately. We
therefore disable $\mathcal{L}_{\mathrm{mam}}$ for the first $10$ epochs
and turn it on with weight $\lambda_{\mathrm{mam}}{=}0.1$ afterwards.

\subsection{Vision-Language Contrastive Alignment}
\label{appx:align}

$\mathcal{L}_{\mathrm{align}}$ pulls the encoder's pre-quantization
continuous latent sequence $h_{1:M}$ toward a frozen Qwen2.5-VL-7B
feature space via InfoNCE at two granularities: a trajectory-level
contrast between mean-pooled action and VL summaries
($\mathcal{L}_{\mathrm{global}}$), and a slot-level contrast between
action and VL slots that are time-aligned within each chunk but
contrasted against all slots across the batch
($\mathcal{L}_{\mathrm{local}}$).

\textbf{Temporal alignment.} The action latent length $M{=}T/r$ and the
VL length $T_v$ generally differ within a chunk. We align them with
\texttt{adaptive\_avg\_pool1d} along the time axis: if $T_v\!>\!M$, the
VL stream is pooled down to $M$; otherwise the action latent is pooled
down to $T_v$. The aligner has no learnable parameters and produces
$\tilde h,\tilde u\!\in\!\mathbb{R}^{B\times M'\times H}$ at a common
length $M'$.

\textbf{InfoNCE implementation details.} Both InfoNCE losses follow the
CLIP recipe~\cite{radford2021learning}: each loss is the symmetric
average of the action$\to$VL direction shown in
Eq.~\ref{eq:global}/\ref{eq:local} and its reverse VL$\to$action
counterpart. The learnable log-scale $s\!\in\!\mathbb{R}$ is initialized
to $\log(1/0.1)$ and clamped above by $\log 100$; the effective scale is
$\gamma{=}\exp(s)$. Features are L2-normalized along the last dimension.
For the slot-level InfoNCE ($\mathcal{L}_{\mathrm{local}}$), action and
VL slots are flattened across the batch into a
$BM'\!\times\!BM'$ logit matrix so that each anchor
$(b,i)\!\in\![B]\!\times\![M']$ is contrasted against all $BM'\!-\!1$
other (chunk, slot) pairs; rows/columns outside the valid-position set
(where either modality is padded) are filled with $-10^9$ before the
softmax and excluded from the cross-entropy targets, so the row-wise CE
only counts valid pairs. For the trajectory-level InfoNCE
($\mathcal{L}_{\mathrm{global}}$), $\bar h$ and $\bar u$ are mask-aware
mean-pools over the valid time positions of each chunk, contrasted
across the batch via a $B\!\times\!B$ logit matrix. The temperature
target is $\kappa{=}0.1$ and the per-direction weights are
$\lambda_{\mathrm{local}}{=}\lambda_{\mathrm{global}}{=}0.25$, which
together with the averaged main-text form
$\mathcal{L}_{\mathrm{align}}{=}\tfrac{1}{2}(\mathcal{L}_{\mathrm{global}}+\mathcal{L}_{\mathrm{local}})$
at $\lambda_{\mathrm{align}}{=}0.5$ (Eq.~\ref{eq:pretrain_loss}) yields a
total alignment contribution of $0.25(\mathcal{L}_{\mathrm{global}}+\mathcal{L}_{\mathrm{local}})$.

\subsection{Next-Frame VL Prediction}
\label{appx:pred}

$\mathcal{L}_{\mathrm{pred}}$ asks the codebook to encode the immediate
physical consequence of the current chunk: a small predictor $G$ takes
the multi-level quantized latent $\tilde{\mathbf{z}}_{1:M}$ and outputs a
vector matching the VL feature of the next frame, with $\ell_1$ loss.

The predictor is a $2$-layer Transformer encoder ($4$ heads, dropout
$0.1$) that processes the latent sequence; we read out the last position
$\tilde{\mathbf{z}}_M^{\mathrm{out}}\!\in\!\mathbb{R}^{H}$ and project it
by $\mathrm{LayerNorm}\!\to\!\mathrm{Linear}(H,H_{\mathrm{vl}})$ to
produce the predicted next-step VL feature in
$\mathbb{R}^{H_{\mathrm{vl}}}$. The prediction target $u_+$ is the next
VL feature after the chunk's VL stream $u_{1:T_v}$ (i.e., the VL feature
at the chunk's next time step in the same view stream), taken from the
same view used during chunk extraction with priority \texttt{face} over
the wrist views; if the \texttt{face} view is missing for a chunk, we
fall back to a randomly chosen view that is present. The loss weight is
$\lambda_{\mathrm{pred}}{=}0.2$.

\subsection{Rotation Geodesic Loss}
\label{appx:geo}

This and the next subsection (App.~\ref{appx:dctsmooth}) cover auxiliary
regularizers that stabilize reconstruction quality without contributing
to codebook semantics.

\textbf{Geodesic loss.} For rotational channels we implement an
$\mathrm{SO}(3)$-aware geodesic constraint:
\begin{equation}
\mathcal{L}_{\mathrm{geo}}
\;=\; \mathbb{E}_{t \in \mathcal{V}_R}
\left[ \frac{1}{\pi}\arccos\!\left(\frac{\mathrm{tr}\!\bigl(R_{t,\mathrm{pred}}^{\top}\, R_{t,\mathrm{gt}}\bigr) - 1}{2}\right) \right],
\label{eq:geo}
\end{equation}
where $\mathcal{V}_R \subseteq \{1, \dots, T\}$ indexes valid rotational
timesteps and $R_{t,\mathrm{pred}}, R_{t,\mathrm{gt}} \in \mathrm{SO}(3)$
are recovered from the 6D representation via
Gram--Schmidt~\cite{hempel20226d,liu2026rdt2}:
\[
b_1=\widehat{r_{1:3}}, \quad
b_2=\widehat{r_{4:6}-(b_1^\top r_{4:6})\,b_1}, \quad
b_3=b_1\times b_2, \quad
R=[b_1\,b_2\,b_3],
\]
where $\widehat{\cdot}$ denotes L2 normalization. The geodesic uses the
numerically stable clamp
$\arccos(\mathrm{clamp}_{(-1+\epsilon,\,1-\epsilon)}(\cdot))$ with
$\epsilon{=}10^{-7}$ to prevent gradient explosion near the boundaries.
The $1/\pi$ prefactor normalizes the angular distance from $[0,\pi]$
radians to $[0,1]$, matching the numerical scale of the translational
$\ell_1$ term in $\mathcal{L}_{\mathrm{rec}}$.

\textbf{Why physical-space evaluation.} If we left the predicted and
target 6D vectors in the normalized $[-1,1]$ range before Gram--Schmidt,
$\arccos(\cdot)$ would be a function of the normalization scale rather
than of an actual angular error: different channels are scaled by
different per-channel ranges, so the recovered ``rotation matrix'' would
not correspond to any physical rotation. We therefore undo the
per-channel MinMax scaling defined in App.~\ref{appx:backbone} on the
rotation 6D segments before recovering
$R_{\mathrm{pred}},R_{\mathrm{gt}}$, so the recovered angle is a true
physical rotation in radians (then scaled to $[0,1]$ by the $1/\pi$
factor of Eq.~\ref{eq:geo}). The loss weight is
$\lambda_{\mathrm{geo}}{=}0.2$.

\subsection{Frequency-Domain and Temporal Smoothness Regularizers}
\label{appx:dctsmooth}

\textbf{Frequency-domain regularizer ($\mathcal{L}_{\mathrm{dct}}$).} A
reconstructed action chunk can have low time-domain $\ell_1$ error per
frame and still contain high-frequency jitter that is invisible to a
frame-wise loss but harmful at deployment time. Following motion-modeling
work that uses DCT to factor a trajectory into a small number of
low-frequency components~\cite{mao2019learning}, we add a DCT-domain
$\ell_1$:
\begin{equation}
\mathcal{L}_{\mathrm{dct}}
\;=\; \bigl\|\,\Phi(\hat x) - \Phi(x)\,\bigr\|_1,
\label{eq:dct}
\end{equation}
where $\Phi$ is the Type-II DCT with orthogonal normalization, implemented
via the standard FFT trick: reorder the input as
$[x_0,x_2,\ldots,x_{n-1},x_{n-2},\ldots,x_3,x_1]$, apply the FFT along
time, multiply by the twiddle factor $\exp(-i\pi k / 2n)$, take the real
part, and rescale by $1/\sqrt{n}$ for $k{=}0$ and $\sqrt{2/n}$ otherwise.
Padding positions are zeroed before the transform. The loss is a
DoF-mask-aware $\ell_1$; the weight is $\lambda_{\mathrm{dct}}{=}0.5$.

\textbf{Temporal smoothness regularizer
($\mathcal{L}_{\mathrm{smooth}}$).} We match the per-frame velocity
(first-order temporal differences) between prediction and target, so
that the reconstructed trajectory reproduces the local dynamics of the
ground truth rather than only its instantaneous positions:
\begin{equation}
\mathcal{L}_{\mathrm{smooth}}
\;=\; \mathbb{E}_{t\in\mathcal{V}_S}\!\left[\bigl\|(\hat x_{t+1}-\hat x_t)-(x_{t+1}-x_t)\bigr\|_1\right],
\label{eq:smooth}
\end{equation}
where $\mathcal{V}_S\!\subseteq\!\{1,\ldots,T{-}1\}$ indexes positions
whose adjacent frames $t, t{+}1$ are both valid under the DoF and padding
masks. Rotation channels are excluded via a static dimension mask: their
first-order difference in the 6D representation is not a clean physical
quantity, and Eq.~\ref{eq:geo} already covers them. The objective
therefore acts on positional, gripper, base-velocity, lift, and head
channels. The weight is $\lambda_{\mathrm{smooth}}{=}0.3$.

\subsection{Training Schedule and Loss Weights}
\label{appx:training}

We pretrain for $100$ epochs on chunks sampled uniformly in
$T\!\in\![8,64]$, with batch size $256$. The optimizer is AdamW (learning
rate $5\!\times\!10^{-5}$, weight decay $0.01$, $\beta{=}(0.9,0.999)$,
gradient clipping $1.0$), under a cosine schedule with $200$-step linear
warm-up and minimum learning rate $10^{-7}$.

The global pretraining objective is
\begin{equation*}
\mathcal{L}_{\mathrm{pre}}
\;=\; \mathcal{L}_{\mathrm{rec}}
\;+\; \lambda_{\mathrm{mam}}\,\mathcal{L}_{\mathrm{mam}}
\;+\; \lambda_{\mathrm{align}}\,\mathcal{L}_{\mathrm{align}}
\;+\; \lambda_{\mathrm{pred}}\,\mathcal{L}_{\mathrm{pred}},
\end{equation*}
where $\mathcal{L}_{\mathrm{rec}}$ aggregates the translational $\ell_1$
loss, the rotation geodesic loss $\mathcal{L}_{\mathrm{geo}}$
(Eq.~\ref{eq:geo}), the VQ commitment loss, and the two stability
regularizers $\mathcal{L}_{\mathrm{dct}}$ (Eq.~\ref{eq:dct}) and
$\mathcal{L}_{\mathrm{smooth}}$ (Eq.~\ref{eq:smooth}). The per-term
weights, grouped by role, are:
\begin{itemize}[topsep=2pt,itemsep=1pt,partopsep=0pt,parsep=1pt,leftmargin=1.5em]
\item \textbf{Semantic supervisions} (App.~\ref{appx:mam},
\ref{appx:align}, \ref{appx:pred}): MAM $\lambda_{\mathrm{mam}}{=}0.1$
(active after the $10$-epoch warm-up of App.~\ref{appx:mam}); VL
contrastive $\lambda_{\mathrm{local}}{=}\lambda_{\mathrm{global}}{=}0.25$
(summing to $\lambda_{\mathrm{align}}{=}0.5$, applied to the averaged
form $\mathcal{L}_{\mathrm{align}}{=}\tfrac{1}{2}(\mathcal{L}_{\mathrm{global}}+\mathcal{L}_{\mathrm{local}})$
of Eq.~\ref{eq:pretrain_loss}) at temperature target $\kappa{=}0.1$;
next-frame VL prediction $\lambda_{\mathrm{pred}}{=}0.2$.
\item \textbf{Reconstruction} (within $\mathcal{L}_{\mathrm{rec}}$;
App.~\ref{appx:backbone}, \ref{appx:geo}): translational $\ell_1$
$\lambda_{\mathrm{l1}}{=}1.0$; rotation geodesic
$\lambda_{\mathrm{geo}}{=}0.2$; VQ commitment $\lambda_{\mathrm{vq}}{=}0.25$.
\item \textbf{Stability regularizers} (within $\mathcal{L}_{\mathrm{rec}}$;
App.~\ref{appx:dctsmooth}): frequency-domain DCT
$\lambda_{\mathrm{dct}}{=}0.5$; temporal smoothness
$\lambda_{\mathrm{smooth}}{=}0.3$.
\end{itemize}

\subsection{Downstream Co-training Objectives}
\label{appx:cotrain}

When X-Tokenizer is used as the discrete supervision interface for a
hybrid discrete-continuous VLA policy (a causal VLM backbone with hidden
states $h_{\mathrm{vlm}}$, co-trained with a continuous Flow Matching
action expert with velocity field $v_\phi$), the downstream co-training
loss combines two terms that together yield the main-text
Eq.~\ref{eq:cotrain}.

With $\mathbf{c}=\{c^{(1)},\dots,c^{(Q)}\}_{1:M}$ extracted by the frozen
X-Tokenizer, the VLM backbone is optimized by autoregressive
cross-entropy over multi-level tokens generated in position-major raster
order (all $Q$ levels at position $i$ before moving to $i{+}1$):
\begin{equation}
\mathcal{L}_{\mathrm{vlm}}
\;=\; -\sum_{i=1}^M \sum_{q=1}^Q
\log p_\psi\!\bigl(c_i^{(q)} \,\big|\, h_{\mathrm{vlm}},\,
c_{<i}^{(1:Q)},\, c_i^{(<q)}\bigr),
\label{eq:vlm_loss}
\end{equation}
where predicting $c_i^{(q)}$ thus conditions on all codes at strictly
earlier positions $c_{<i}^{(1:Q)}$ and on the outer-level codes at the
same position $c_i^{(<q)}$ (teacher-forcing). The Flow Matching expert is
conditioned on $h_{\mathrm{vlm}}$ to regress continuous trajectories
$x_{1:T}$ at a randomly sampled time $t\!\in\![0,1]$:
\begin{equation}
\mathcal{L}_{\mathrm{fm}}
\;=\; \mathbb{E}_{t, x_t}
\left[ \bigl\| v_\phi(x_t, t \mid h_{\mathrm{vlm}}) - u^{\star}_t \bigr\|^2_2 \right],
\label{eq:fm_loss}
\end{equation}
where $u^{\star}_t$ is the target Flow Matching velocity. The two terms
are combined with relative weight $\lambda_{\mathrm{fm}}$ to form the
co-training loss $\mathcal{L}_{\mathrm{vlm}} + \lambda_{\mathrm{fm}}\,
\mathcal{L}_{\mathrm{fm}}$ (i.e., the right-hand side of
Eq.~\ref{eq:cotrain}).

\section{Pretraining Datasets and Embodiments}
\label{appx:data}

\subsection{Pretraining Corpus}
\label{appx:corpus}

The pretraining mixture is assembled from X2Robot-internal data, public
academic datasets, and third-party releases, all converted to the $26$-D
delta-action layout of App.~\ref{appx:backbone}.
Tab.~\ref{tab:pretrain_corpus} lists the sources we draw from.
Trajectories from different sources are not reweighted at sampling
time---each chunk is sampled uniformly from the union after
filtering---so high-volume sources carry proportionally more weight in
the gradient. For every source we drop trajectories with fewer than
$T_{\min}{=}8$ valid action frames, missing all camera views, or
corrupted action streams (NaN/out-of-range values, broken
proprio--control alignment); per-channel normalization quantiles
($q_{0.1\%}$ and $q_{99.9\%}$) are computed per \texttt{robot\_type} on
the raw streams; chunks are extracted at chunk length sampled uniformly
in $[T_{\min}, T_{\max}]$ with $T_{\max}{=}64$, anchored to the
chunk-level reference $o$ defined in App.~\ref{appx:backbone}. After
filtering, the union contains $\sim\!2.4$M trajectories and
$\sim\!2.0$B valid action frames.

\begin{table}[H]
\centering
\caption{\small \textbf{Pretraining corpus.} Source datasets grouped by
provenance; ``\#Robots'' counts the distinct \texttt{robot\_type} entries
with data in the current mixture.}
\label{tab:pretrain_corpus}
\footnotesize
\renewcommand{\arraystretch}{1.15}
\setlength{\tabcolsep}{4pt}
\begin{tabular}{@{}p{2.6cm} p{5.6cm} c l@{}}
\toprule
\textbf{Source} & \textbf{Description} & \textbf{\#Robots} & \textbf{Ref.} \\
\midrule
\multicolumn{4}{@{}l}{\textit{X2Robot internal}} \\
\midrule
X2Robot (in-house) & --- & $11$ & \cite{zhai2025igniting} \\
\midrule
\multicolumn{4}{@{}l}{\textit{Large cross-embodiment academic releases}} \\
\midrule
AgiBotWorld   & humanoid dual-arm manipulation                                      & $1$  & \cite{bu2025agibot_iros, agibotworld2026} \\
DROID         & in-the-wild Franka tabletop                                         & $1$  & \cite{khazatsky2024droid} \\
RoboTwin~2.0  & sim, dual-arm benchmark (Aloha/Arx5/Franka/Piper/UR5)               & $5$  & \cite{chen2025robotwin} \\
\midrule
\multicolumn{4}{@{}l}{\textit{Multi-platform third-party corpora}} \\
\midrule
RoboMind/V2   & Franka/UR5/Agilex/Ark (V1+V2)                                       & $11$ & \cite{wu2024robomind, hou2025robomind} \\
RoboCoin      & bimanual collection (Cobot/Aloha/Alpha-bot-2/MMK2/Leju/Realman/A2D) & $7$  & \cite{wu2025robocoin} \\
RoboChallenge & competition tasks (Franka/UR5/Aloha/Arx5)                           & $4$  & \cite{robochallenge2025} \\
R1Lite        & Galaxea R1-Lite humanoid                                            & $1$  & \cite{jiang2025galaxea} \\
RealOmni      & open multi-modal robot dataset                                      & $1$  & \cite{realomni2025} \\
\midrule
\multicolumn{4}{@{}l}{\textit{Single-embodiment / per-task releases}} \\
\midrule
Bridge-V2      & WidowX low-cost manipulation                                       & $1$ & \cite{walke2023bridgedata} \\
Fractal-RT     & Google EveryDay Robot (RT-1 corpus)                                & $1$ & \cite{brohan2022rt1} \\
BC-Z           & Google EveryDay Robot, BC-Z                                        & $1$ & \cite{jang2022bc} \\
FurnitureBench & long-horizon assembly (Franka)                                     & $1$ & \cite{heo2025furniturebench} \\
Open-X subsets & Stanford (hydra, kuka-multimodal); UT-Austin (buds, sailor, sirius, mutex); Berkeley (autolab-ur5, cable-routing, fanuc-manipulation) & $9$ & \cite{o2024open} \\
\midrule
\multicolumn{2}{@{}r}{\textbf{Total}} & $\mathbf{54}$ & \\
\bottomrule
\end{tabular}
\end{table}

\subsection{Embodiment Coverage}
\label{appx:embodiments}

X-Tokenizer is pretrained on data from $54$ \texttt{robot\_type} entries
drawn from the corpora listed in App.~\ref{appx:corpus}. These $54$
\texttt{robot\_type}s are mapped to $17$ arm families by the hardware
lookup used in our cross-embodiment analyses;
Tab.~\ref{tab:embodiment_coverage} lists the $17$ families, grouped by
manipulator type. Our \texttt{robot\_types} registry additionally
declares roughly $15$ further embodiments that share the same
delta-action layout (App.~\ref{appx:backbone}) but are not yet present
in the current training mixture, for a total of more than $70$
predefined embodiments.

\begin{table}[H]
\centering
\caption{\small \textbf{The $17$ arm families} represented in the
X-Tokenizer pretraining corpus, grouped by manipulator type. Each
\texttt{robot\_type} in our registry is mapped to exactly one of these
$17$ families.}
\label{tab:embodiment_coverage}
\renewcommand{\arraystretch}{1.2}
\begin{tabular}{@{}>{\bfseries}l p{8.5cm}@{}}
\toprule
Family & Description \\
\midrule
\multicolumn{2}{@{}l}{\textit{Single-arm collaborative manipulators}} \\
\midrule
Franka      & Franka Emika Panda 7-DoF arm. \\
UR5         & Universal Robots UR5 6-DoF collaborative arm. \\
\midrule
\multicolumn{2}{@{}l}{\textit{Lightweight research / low-cost arms}} \\
\midrule
Piper       & Realman ultra-light biomimetic arm. \\
ViperX      & Trossen Robotics Dynamixel-based arm (Aloha series). \\
ARX5        & ARX 5--6 DoF arm. \\
WidowX      & Trossen Robotics compact arm. \\
X2Arm       & X2Robot internal arm~\cite{zhai2025igniting}. \\
\midrule
\multicolumn{2}{@{}l}{\textit{High-DoF research arms}} \\
\midrule
Realman     & Realman 7-DoF series. \\
Ark         & Ark arm series. \\
\midrule
\multicolumn{2}{@{}l}{\textit{Humanoid and mobile platforms}} \\
\midrule
AgiBot      & AgiBot (Yuanzheng) humanoid robot (dual-arm). \\
Leju        & Leju (Kurui) humanoid robot. \\
R1Lite      & Galaxea R1-Lite humanoid. \\
AlphaBot    & RoboCoin Alpha Bot 2 dual-arm platform. \\
MMK2        & Composite mobile manipulator. \\
GoogleRobot & Google mobile manipulation platform (RT-1 / RT-2). \\
\midrule
\multicolumn{2}{@{}l}{\textit{Specialized devices}} \\
\midrule
UMI         & Universal Manipulation Interface (handheld gripper data). \\
A2D         & RoboCoin RuanTong A2D specialized arm. \\
\bottomrule
\end{tabular}
\end{table}

\subsection{Baseline Tokenizer Implementation}
\label{appx:baselines}

The two learned baselines are loaded from their public checkpoints
without retraining or per-channel re-fitting, so the comparison reflects
the same artifacts already deployed by the community.
\textbf{FAST}~\cite{pertsch2025fast} is loaded via the Hugging Face
\texttt{AutoProcessor} interface; encoding maps a normalized action
chunk in $[-1, 1]$ to a variable-length integer sequence, and decoding
inverts it back to the original $T\!\times\!D$ shape. The
\textbf{RDT2~VQ}~\cite{liu2026rdt2} single-codebook VQ-VAE is used only
in the noise probe of \S\ref{sec:exp:tokenizer:noise} as a control for
``what does a single-level VQ buy you under noise'' and is not part of
the downstream ablation. The non-learned \textbf{$256$-bin per-channel
uniform quantizer} partitions each channel of the normalized action
into $256$ equal bins on $[-1, 1]$ and maps each value to its bin
centre; it carries no learned structure and only anchors the
reconstruction-$\ell_1$ axis in \S\ref{sec:exp:tokenizer:ablation}.
Other learned tokenizers are not directly compared in our protocol:
\textbf{VQ-BeT}~\cite{lee2024behavior} has no released cross-embodiment
checkpoint at the scale of FAST, so a head-to-head run would conflate
``tokenizer'' with ``training corpus'';
\textbf{FASTer}~\cite{liu2025faster} has not released public weights at
the time of writing; and
\textbf{ActionCodec}~\cite{dong2026actioncodec} targets the
purely-discrete autoregressive setting structurally incompatible with
our mixed discrete-continuous co-training (\S\ref{sec:exp:wallx}).

\section{Downstream Training Configurations}
\label{appx:downstream}

\subsection{RoboTwin~2.0 Training}
\label{appx:robotwin}

We start from a publicly released
Wall-OSS~\cite{zhai2025igniting} VLA policy checkpoint with full
degrees of freedom (dual arms, base, lift, head) trained for $400$k
pretraining steps. The frozen X-Tokenizer is attached to this backbone
in place of the discrete action interface; all other modules of
Wall-OSS are inherited from the public checkpoint without modification.
On the standard $50$ dual-arm task suite, each task contributes $50$
Clean and $500$ Randomized demonstrations, for a total of
$\sim\!27.5$k trajectories; the cross-embodiment co-training
experiment (Agilex / Arx5 / Franka / Piper / UR5) uses the same
per-task counts shared across the five embodiments. We fine-tune the
full system for $70$k steps with global batch size $128$, AdamW (lr
$5{\times}10^{-5}$, weight decay $0.01$, $\beta{=}(0.9, 0.999)$,
gradient clipping $1.0$), and a cosine learning-rate schedule with
$200$-step linear warm-up and minimum learning rate $10^{-7}$; action
chunks are sampled with the same length distribution as the
X-Tokenizer pretraining ($T\!\in\![8, 64]$). Each task is rolled out
$100$ times under Easy (Clean) and Hard (Randomized, with strong
domain randomization over background clutter, lighting, table height,
and distractor objects); the per-task success rate is averaged over
the $50$ tasks for the Avg column.

\begin{longtable}{@{}p{6.2cm}cc@{}}
\caption{\small \textbf{Per-task RoboTwin~2.0 success rates} for the
Wall-OSS+X-Tokenizer run in Fig.~\ref{fig:robotwin_dualarm}. Each entry
is measured over $100$ rollouts.}
\label{tab:robotwin_per_task}\\
\toprule
\textbf{Task} & \textbf{Easy} & \textbf{Hard} \\
\midrule
\endfirsthead
\toprule
\textbf{Task} & \textbf{Easy} & \textbf{Hard} \\
\midrule
\endhead
\midrule
\multicolumn{3}{r}{\footnotesize Continued on next page} \\
\endfoot
\bottomrule
\endlastfoot
\texttt{adjust\_bottle} & $100.00\%$ & $100.00\%$ \\
\texttt{beat\_block\_hammer} & $88.00\%$ & $78.00\%$ \\
\texttt{blocks\_ranking\_rgb} & $86.00\%$ & $90.00\%$ \\
\texttt{blocks\_ranking\_size} & $46.00\%$ & $46.00\%$ \\
\texttt{click\_alarmclock} & $80.00\%$ & $88.00\%$ \\
\texttt{click\_bell} & $90.00\%$ & $88.00\%$ \\
\texttt{dump\_bin\_bigbin} & $91.00\%$ & $93.00\%$ \\
\texttt{grab\_roller} & $100.00\%$ & $100.00\%$ \\
\texttt{handover\_block} & $81.00\%$ & $76.00\%$ \\
\texttt{handover\_mic} & $94.00\%$ & $92.00\%$ \\
\texttt{hanging\_mug} & $31.00\%$ & $20.00\%$ \\
\texttt{lift\_pot} & $90.00\%$ & $92.00\%$ \\
\texttt{move\_can\_pot} & $96.00\%$ & $100.00\%$ \\
\texttt{move\_pillbottle\_pad} & $96.00\%$ & $87.00\%$ \\
\texttt{move\_playingcard\_away} & $97.00\%$ & $90.00\%$ \\
\texttt{move\_stapler\_pad} & $89.00\%$ & $78.00\%$ \\
\texttt{open\_laptop} & $96.00\%$ & $86.00\%$ \\
\texttt{open\_microwave} & $69.00\%$ & $60.00\%$ \\
\texttt{pick\_diverse\_bottles} & $82.00\%$ & $65.00\%$ \\
\texttt{pick\_dual\_bottles} & $94.00\%$ & $73.00\%$ \\
\texttt{place\_a2b\_left} & $85.00\%$ & $77.00\%$ \\
\texttt{place\_a2b\_right} & $84.00\%$ & $80.00\%$ \\
\texttt{place\_bread\_basket} & $77.00\%$ & $83.00\%$ \\
\texttt{place\_bread\_skillet} & $82.00\%$ & $84.00\%$ \\
\texttt{place\_burger\_fries} & $94.00\%$ & $96.00\%$ \\
\texttt{place\_can\_basket} & $84.00\%$ & $72.00\%$ \\
\texttt{place\_cans\_plasticbox} & $99.00\%$ & $97.00\%$ \\
\texttt{place\_container\_plate} & $97.00\%$ & $97.00\%$ \\
\texttt{place\_dual\_shoes} & $93.00\%$ & $87.00\%$ \\
\texttt{place\_empty\_cup} & $100.00\%$ & $99.00\%$ \\
\texttt{place\_fan} & $86.00\%$ & $80.00\%$ \\
\texttt{place\_mouse\_pad} & $56.00\%$ & $53.00\%$ \\
\texttt{place\_object\_basket} & $95.00\%$ & $80.00\%$ \\
\texttt{place\_object\_scale} & $79.00\%$ & $69.00\%$ \\
\texttt{place\_object\_stand} & $95.00\%$ & $85.00\%$ \\
\texttt{place\_phone\_stand} & $82.00\%$ & $74.00\%$ \\
\texttt{place\_shoe} & $97.00\%$ & $94.00\%$ \\
\texttt{press\_stapler} & $93.00\%$ & $92.00\%$ \\
\texttt{put\_bottles\_dustbin} & $58.00\%$ & $76.00\%$ \\
\texttt{put\_object\_cabinet} & $73.00\%$ & $77.00\%$ \\
\texttt{rotate\_qrcode} & $89.00\%$ & $84.00\%$ \\
\texttt{scan\_object} & $77.00\%$ & $72.00\%$ \\
\texttt{shake\_bottle} & $100.00\%$ & $100.00\%$ \\
\texttt{shake\_bottle\_horizontally} & $100.00\%$ & $100.00\%$ \\
\texttt{stack\_blocks\_three} & $87.00\%$ & $87.00\%$ \\
\texttt{stack\_blocks\_two} & $97.00\%$ & $94.00\%$ \\
\texttt{stack\_bowls\_three} & $81.00\%$ & $61.00\%$ \\
\texttt{stack\_bowls\_two} & $92.00\%$ & $88.00\%$ \\
\texttt{stamp\_seal} & $62.00\%$ & $66.00\%$ \\
\texttt{turn\_switch} & $45.00\%$ & $38.00\%$ \\
\midrule
\textbf{Average} & $\mathbf{84.70\%}$ & $\mathbf{80.88\%}$ \\
\end{longtable}

\subsection{Real-World Training and Evaluation}
\label{appx:wallx}

\textbf{Data and training schedule.} Real-robot supervision is
collected on the $7$ tabletop tasks of
Fig.~\ref{fig:real_world_combined} with $\sim\!500$ teleoperated
trajectories per task, totalling $\sim\!3.5$k trajectories; the $7$
tasks cover both short-horizon manipulation (e.g., pick-up-cup,
push-towel, place-tape) and long-horizon reasoning (arrange-flowers,
turn-on-light-switch). Action data are mixed with $\sim\!480$k
multimodal grounding samples so that grounding samples occupy $25\%$
of each batch, fixed throughout training; the same $480$k samples are
used across all four variants and are held out from the VQA
evaluation set. Grounding samples cover four sub-types at comparable
scale: (i) point grounding, (ii) bounding-box grounding, (iii)
end-effector grounding, and (iv) trajectory grounding chunks.
Training uses AdamW (lr $1{\times}10^{-4}$, weight decay $0.01$,
$\beta{=}(0.9, 0.999)$, gradient clipping $1.0$) for $500$k steps with
batch size $8$ per GPU and gradient accumulation $2$ (effective batch
size $16$/GPU), under a $500$-step linear warm-up and cosine decay to
minimum learning rate $10^{-6}$.

\textbf{Evaluation protocol.} Real-robot tasks are each rolled out
$10$ times on the physical platform; per-task progress rate (PR) is
human-scored stage-by-stage on the rubric of App.~\ref{appx:scoring},
with absolute values systematically lower than binary success rates
because the rubric awards credit for partial completions. VQA
grounding accuracy is evaluated on
\textsc{x2-grounding-point-object} ($N{=}107$ held-out samples from
our internal point-grounding annotations), with a prediction counted
correct iff the model's predicted $\langle x,y\rangle$ point falls
inside the ground-truth segmentation mask.
\paragraph{Interpreting the real-world ablation.}
The four action-interface variants in Fig.~\ref{fig:real_world_combined}
separate several effects. First, replacing the original continuous flow
interface with FAST introduces a token-level action interface, which is
better matched to the language-token training signal used by the VLM
backbone and coincides with a large VQA gain. We therefore view the
Wall-OSS$\to$FAST jump mainly as an interface-level effect rather than
as evidence about X-Tokenizer specifically.

Second, FAST$\to$RVQ (no-aux) isolates the effect of a multi-level
discrete codebook without semantic auxiliary heads. This improves VQA
($75.7\!\to\!79.4$), suggesting that the hierarchy provides a richer
discrete supervision signal to the backbone, but it lowers manipulation
and long-horizon PR. Thus reconstruction-only RVQ improves the
representation side but is not sufficient for action quality.

Third, adding MAM, Align, and Pred yields the full X-Tokenizer. This
raises VQA to $85.9\%$, increases the short-horizon manipulation
aggregate from $73.0$ to $80.6$, and improves long-horizon PR from
$59.5$ to $69.25$ relative to RVQ (no-aux). These gains are consistent
with the intended role of semantic supervision: it turns the multi-level
codebook into an action interface that is easier for the VLM backbone to
use.

Finally, the main per-task regression appears on
\emph{distribute-blocks-by-color}, a repetitive placement task where
stage credit depends heavily on low-level placement accuracy. This is
consistent with the reconstruction trade-off in
\S\ref{sec:exp:tokenizer:ablation}, but the aggregate manipulation and
long-horizon results remain higher for X-Tokenizer.

\begin{table}[H]
\centering
\caption{\small \textbf{Step-wise scoring rules} for the real-world
tasks of \S\ref{sec:exp:wallx} (max $10$ pts per task; PR =
accumulated score / $10$, averaged over $10$ rollouts, on $0$--$100$
scale).}
\label{tab:scoring_rules}
\renewcommand{\arraystretch}{1.3}
\begin{tabular}{@{} >{\bfseries}l p{8.5cm} @{}}
\toprule
Task Name & \textbf{Scoring Rule (Max 10 pts)} \\
\midrule
\multicolumn{2}{@{}l}{\textit{Manipulation Tasks}} \\
\midrule
Pick up cup                  & Push plate (3), upright cup (2), pick up (2), place (2), retract (1) \\
Push towel                   & Successfully push the red long towel onto the target position (10) \\
Distribute blocks (by color) & $3$ pts per correct block placement; $3$ blocks (9), retract (1) \\
Stack bottle                 & Fully nest one bottle into another (10) \\
Place tape                   & Grasp the wide tape (5) + place into the plate (5) \\
\midrule
\multicolumn{2}{@{}l}{\textit{Long-Horizon Reasoning Tasks}} \\
\midrule
Arrange flowers              & Per flower: grasp (1.5) + place into vase (1.5); $3$ flowers (9), retract (1) \\
Turn on light switch         & Move to switch position (3), press switch (4), retract (3) \\
\bottomrule
\end{tabular}
\end{table}

\section{Scoring Rubric for Real-World Tasks}
\label{appx:scoring}

We follow the per-task progress-rate (PR) scoring protocol introduced
in the Wall-OSS technical report~\cite{zhai2025igniting}: each task
is decomposed into key manipulation stages with stage-wise partial
credit summing to a maximum of $10$ points, and the PR for one rollout
is the accumulated score divided by $10$ (reported on a $0$--$100$
scale in our main tables). Compared with binary success/failure, this
rubric makes the metric sensitive to \emph{where} a policy fails---%
e.g., whether the cup is reachable, gripped, lifted, but dropped
during placement---which is essential for diagnosing long-horizon
reasoning tasks where partial completion is common. The full design
rationale and per-stage credit philosophy are documented
in~\cite{zhai2025igniting}; we adopt the same protocol verbatim and
list our per-task rules in Tab.~\ref{tab:scoring_rules}.

\end{document}